\PassOptionsToPackage{table}{xcolor}
\documentclass[manuscript,screen]{acmart}
\usepackage[table]{xcolor}

\AtBeginDocument{%
  \providecommand\BibTeX{{%
    \normalfont B\kern-0.5em{\scshape i\kern-0.25em b}\kern-0.8em\TeX}}}

\usepackage{balance}  
\usepackage{graphics} 

\usepackage{adjustbox}

\usepackage{url}
\usepackage{fancybox}
\usepackage{multirow}
\usepackage{flushend}
\usepackage{booktabs}
\usepackage{tabularx}
\usepackage{comment}
\usepackage{array}
\usepackage[flushleft]{threeparttable}
\usepackage{mdframed}
\graphicspath{{}{images/}{dia/}}
\DeclareGraphicsExtensions{.pdf,.png}

\usepackage{listings}

\usepackage{courier}
\usepackage{hyperref}

\usepackage{xpatch}

\xpatchcmd{\refstepcounter}{%
  \stepcounter{#1}%
}{%
  \stepcounter{#1}%
  \xdef\scr{\number\value{#1}}%
}{\typeout{success}}{\typeout{failure}}

\newcounter{o}
\usepackage{soul}
\usepackage{tikz}
\definecolor{1c1}{RGB}{188,162,6}
\definecolor{1c2}{RGB}{137,129,80}
\definecolor{1c3}{RGB}{239,167,31}
\definecolor{1c4}{RGB}{88,194,241}
\definecolor{1c5}{RGB}{6,180,188}

\tikzset{mynode/.style={draw=white,solid,circle,fill=green,inner sep=1pt, thick,
text=black}}
\tikzset{arrow line/.style={dashed, line width= 2.5pt, color=#1}}

\usepackage{tcolorbox}

\tcbuselibrary{skins}

\newtcolorbox{rqbox}[1]
{
  before skip=1em, 
  after skip=1em, 
  colframe = black!60,
  colback  = black!10,
  coltitle = black!90,  
  title    = \textbf{#1},
  enhanced,
  attach boxed title to top center={yshift=-3mm,yshifttext=-1mm},
  boxed title style={size=small, colback=black!30}
}

\def\bf{\textbf}

\def\fig {Figure~}

\def\tbl {Table~}

\def\sec {Section~}

\def\it{\textit}

\def\tt{\mct}
\newcommand{\ib}[1]{{\textbf {\textit { #1}}}}

\newcommand{\mct}[1]{{\footnotesize {\texttt {#1}}}}

\usepackage{paralist}

\newcommand{\nd}{\vspace{1mm}\noindent}
\usepackage{caption}
\usepackage{tikz}

\lstdefinestyle{inlinecode}{basicstyle={\ttfamily\scriptsize\bfseries}}

\newcommand{\urls}[1]{{\scriptsize\url{#1}}}
\usepackage{tcolorbox}

\newcommand{\bigCI}{\mathrel{\text{\scalebox{1.07}{$\perp\mkern-10mu\perp$}}}}

\usepackage{paralist}
\usepackage[outercaption]{sidecap}
\usepackage [autostyle, english = american]{csquotes}
\MakeOuterQuote{"}
\newcounter{scn}
\usepackage[shortlabels]{enumitem}
\usepackage{tikz}
\usepackage{styles/pgf-pie}
\usetikzlibrary{positioning,shadows}
\usepackage{balance}

\newif\ifpienumberinlegend
\pgfkeys{/number in legend/.code=
    \expandafter\let\expandafter\ifpienumberinlegend
    \csname if#1\endcsname
    \ifpienumberinlegend

    \def\beforenumber##1\afternumber{}%
    \fi,
    /number in legend/.default=true
}
\definecolor{1c1}{RGB}{188,162,6}
\definecolor{1c2}{RGB}{137,129,80}
\definecolor{1c3}{RGB}{239,167,31}
\definecolor{1c4}{RGB}{88,194,241}
\definecolor{1c5}{RGB}{6,180,188}

\tikzset{mynode/.style={draw=white,solid,circle,fill=green,inner sep=1pt, thick,
text=black}}
\tikzset{arrow line/.style={dashed, line width= 2.5pt, color=#1}}

\usepackage{bchart}

\definecolor{ao(english)}{rgb}{0.0, 0.5, 0.0}
\def\test#1{%
\ifnum0#1>0
      #1\%
    \fi
}

\let\oldquote\quote
\let\endoldquote\endquote
\renewenvironment{quote}[2][]
  {\if\relax\detokenize{#1}\relax
     \def\quoteauthor{#2}%
   \else
     \def\quoteauthor{#2~---~#1}%
   \fi
   \oldquote}
  {\par\nobreak\smallskip\hfill(\quoteauthor)%
   \endoldquote\addvspace{\bigskipamount}}
\usepackage[noorphans,vskip=0.5ex]{quoting}

\usepackage{subfig}

\hypersetup{
    colorlinks=true,
    linkcolor=blue,
    filecolor=magenta,      
    urlcolor=cyan,
}

\begin{document}

\title{Applications and Challenges of Fairness APIs in Machine Learning Software}

\author{Ajoy Das}
\email{ajoy.das@ucalgary.ca}
\affiliation{
    \institution{DISA Lab, University of Calgary} 
    \country{Canada}
 } 
\author{Gias Uddin}
\email{guddin@yorku.ca}
\affiliation{
    \institution{DISA Lab, York University and University of Calgary} 
    \country{Canada}
 }

 \author{Shaiful Chowdhury}
\email{shaiful.chowdhury@umanitoba.ca}
\affiliation{
    \institution{University of Manitoba} 
    \country{Canada}
 }
\author{Mostafijur Rahman Akhond}
\email{mostafij@yorku.ca}
\affiliation{
    \institution{DISA Lab, York University} 
    \country{Canada}
 }
 
\author{Hadi Hemmati}
\email{hadi.hemmati@ucalgary.ca}
\affiliation{
    \institution{SEA Lab, University of Calgary} 
    \country{Canada}
 }


\begin{abstract}
Machine Learning software systems are frequently used in our day-to-day lives. Some of these systems are used in various sensitive environments to make life-changing decisions. Therefore, it is crucial to ensure that these AI/ML systems do not make any discriminatory decisions for any specific groups or populations. In that vein, different bias detection and mitigation open-source software libraries (aka API libraries) are being developed and used. In this paper, we conduct a qualitative study to understand in what scenarios these open-source fairness APIs are used in the wild, how they are used, and what challenges the developers of these APIs face while developing and adopting these libraries. We have analyzed 204 GitHub repositories (from a list of 1885 candidate repositories) which used 13 APIs that are developed to address bias in ML software. We found that these APIs are used for two primary purposes (i.e., learning and solving real-world problems), targeting 17 unique use-cases. Our study suggests that developers are not well-versed in bias detection and mitigation; they face lots of troubleshooting issues, and frequently ask for opinions and resources. Our findings can be instrumental for future bias-related software engineering research, and for guiding educators in developing more state-of-the-art curricula.

\end{abstract}

\ccsdesc[500]{Computing methodologies~Machine learning}

\keywords{bias, api, github, fairness}
\maketitle

%

%

\section{Introduction}\label{sec:intro}
Fairness is the absence of any prejudice or favoritism toward an individual or
group based on their inherent or acquired characteristics
\cite{mehrabiSurveyBiasFairness2021}. The absence of fairness leads to bias,
which may affect us deeply. Researchers have found different types of
biases in recommendation systems \cite{bhadaniBiasesRecommendationSystem2021},
image recognition systems \cite{schwemmerDiagnosingGenderBias2020}, natural
language processing systems \cite{blodgettLanguageTechnologyPower2020},
recruiting systems \cite{AmazonScrapsSecret}, loan and credit approval systems
\cite{chenFairLendingNeeds2018}, and even health care systems
\cite{vokingerMitigatingBiasMachine2021}, just to name a few. 
For example, a study by 
Angwin \textit{et al.}~\cite{mattujefflarsonMachineBias} showed that an
algorithm that was used by different prisons in the US to predict future
criminals were twice as likely to label black defenders as high-risk who
eventually did not re-offend as compared to white defenders. Likewise,
\emph{Google Translate}, the most popular translation engine in the world, shows
bias on the basis of gender in a sentence. For example, ``She is an engineer, he
is a nurse'' is translated into Turkish and then again into English becomes ``He
is an engineer, she is a nurse''
\cite{caliskanSemanticsDerivedAutomatically2017}.

Given the importance of reducing bias, ensuring \emph{Fairness} has become an actively
researched topic in Machine Learning (ML) systems. Researchers are  developing
novel methods for
detecting~\cite{hardtEqualityOpportunitySupervised2016a,gajaneFormalizingFairnessPrediction2018,vermaFairnessDefinitionsExplained2018,yaoParityFairnessObjectives2017,pessachAlgorithmicFairness2020},
and mitigating bias by different pre-processing
\cite{calmonOptimizedPreProcessingDiscrimination2017,feldmanCertifyingRemovingDisparate2015},
in-processing
\cite{celisClassificationFairnessConstraints2020,kamishimaFairnessawareClassifierPrejudice2012,zhangMitigatingUnwantedBiases2018},
and post-processing
\cite{hardtEqualityOpportunitySupervised2016a,pleissFairnessCalibration2017,kamiranDecisionTheoryDiscriminationAware2012}
methods. The Software Engineering (SE) community has focused on understanding bias in
different SE scenarios
\cite{paulExpressionsSentimentsCode2019}, and designing automated approach to
detect and mitigate
bias~\cite{chakrabortyBiasMachineLearning2021,zhouBiasHealOntheFlyBlackBox2021,yangBiasRVUncoveringBiased2021,hortFaireaModelBehaviour2017}.
Consequently, open-source software libraries (aka API libraries) like Fairlearn~\cite{FairlearnPythonPackage2018},
and AI Fairness 360~\cite{AIFairness3602018}, are being developed that focus on
bias detection and algorithms. APIs (Application Programming Interfaces) aka software libraries 
are interfaces to reusable software components. While the concept of APIs can be broad, 
in this paper, we denote APIs as functionalities provided by software libraries whose features we can reuse in our source code without 
reimplementing the features. These APIs can
assist the ML developers and the ML research community to incorporate bias
testing and mitigating approaches easily into their ML software development pipeline. 

There is no study that focuses on how different open-source bias detection
and mitigation APIs are used in the real world, and what frequent issues
the ML developers face while developing and using these APIs. In this paper, we aim to 
understand the use cases and challenges of the application of open-source fairness APIs. Our
objectives are two folds:
\begin{enumerate}
  \item to help the ML and SE  research community to understand the
importance and challenges of this field for conducting better-guided research
studies, and
\item to help the educators to plan and build relevant state-of-the-art
curricula for future ML developers.
\end{enumerate} 



We have two major phases in our study. 
In phase 1, we investigate how our selected list of 13
fairness APIs are used in open-source ML software by answering three
research questions (RQ1 - RQ3). For this, we found 204 repositories in GitHub 
that truly use at least one of the
selected fairness APIs. 
These repositories contain use-cases for both generic
learning and exploration as well as solving real-world problems. The use-cases
solving non-generic purpose consist of three different activity types
(i.e., Analysis, Prediction, and Operation). We also found that these
non-generic repositories used only two categories of detection metrics and 13
different types of mitigation algorithms. We also observe an increasing use of
detection and mitigation approaches over time.


Then, in phase 2, we investigate what challenges the developers face while
developing and using the API libraries by answering one research question (RQ4).
For this, we collected 4212 issue discussions (989 issues + 3223 issue
comments), from the GitHub repositories of the 13 API libraries, where the API
developers and API users report various issues related to these API
libraries.
After applying topic modeling to these issue discussions, we found 10 different
topics related to six different stages of the software development life cycle
(SDLC).
Most of the issue discussions are related to queries regarding requirement
analysis, deployment, and maintenance phases. In general, API users lack expertise about how to use the fairness APIs effectively and how to validate
their outcomes after applying the mitigation approaches.



The findings of this paper can be summarized as follows.
\begin{description}[leftmargin=30pt]

\item[\bf{RQ1}] \bf{What use cases the fairness APIs are applied for in the open-source Machine Learning Software Applications (MLSAs)?} The applications of bias detection and mitigation APIs are
diverse and include extremely sensitive use-cases such as decision-making in legal, business, and healthcare systems.
\item[\bf{RQ2}] \bf{How are the biases detected in these
  use-cases?} For bias detection in ML systems, the development community
focuses more on measuring group fairness than individual or subgroup fairness.
\item[\bf{RQ3}] \bf{How are the biases mitigated in these use-cases?} 
For bias mitigation, in-processing is the most commonly used approach, where Fairness Constraints techniques are used to limit the impact of bias based on sensitive attributes (e.g., gender) putting various types of fairness-related constraints within the trained ML model.
\item[\bf{RQ4}] \bf{What are the topics found in the issues that developers reported while using the APIs?} Developing fairness APIs or even applying them is
challenging for ML developers.
They struggle to understand what \emph{fairness} means in a given context.
They also frequently discuss bugs and installation issues and often ask
for opinions about what specific methodology they should follow while adopting a
fairness API.
\end{description}

Our results are primarily based on qualitative analysis. 
The findings from our study can guide several stakeholders. ML researchers can learn from the real-world usage of the fairness APIs, which can inform them of the potential opportunities to 
further advance the field of fairness API design and development by focusing on specific use cases. 
SE practitioners and researchers can learn from the challenges of the users and developers of the API libraries and develop techniques and tools to improve the quality assurance and 
adoption of the API libraries. Fairness API developers and ML educators can join hands to develop better documentation 
for their API libraries so that users can easily define and evaluate fairness constraints. 

\nd\bf{Replication Package} containing the data files can be found at \url{https://github.com/disa-lab/bias_api_study}

\section{Background}\label{sec:background}
In this section, we briefly overview the major concepts behind bias detection (in \sec\ref{sec:background-bias-detection}) and mitigation (in \sec\ref{sec:background-bias-mitigation}) approaches, based on which we conducted the analysis in our study. 

\subsection{Bias Detection Approaches}\label{sec:background-bias-detection}
A suite of metrics are proposed in ML literature to detect biases in ML software.
Our classification of the metrics is primarily based on the fairness metrics
classification suggested by Castelnovoa \textit{et
al.}~\cite{castelnovoZooFairnessMetrics2021}. However, we found that the classification of Castelnovoa
\textit{et al.} neither included fairness metrics that can handle multiple
sensitive attributes, nor the fairness metrics that can simultaneously be
applied to measure both group and individual fairness (e.g., entropy indices
\cite{speicherUnifiedApproachQuantifying2018}). Therefore, we also include these
other types of fairness metrics in the classification, which ultimately resulted in 
five different classes of
bias detection metrics: \begin{inparaenum}[(1)]
\item Group,
\item Group-Individual,
\item Individual,
\item Subgroup, and 
\item Utility fairness metrics.
\end{inparaenum} We discuss below the five types of fairness metrics with examples. 

\nd\bf{\ul{Group fairness metrics}} are used to detect bias in the treatment that different groups 
receive from an ML model's prediction. Group fairness is divided into three subcategories: 

\begin{enumerate}[leftmargin=10pt]
\item \ib{Independence} type considers that sensitive characteristics are
statistically independent of the model's prediction. 
These types of metrics include demographic
parity and statistical parity \cite{dworkFairnessAwareness2011} that do not use
target variables during their fairness measurement. Here, a target or real value is the dependent variable of a dataset that we want to predict or understand using machine learning techniques.

Consider students
from two different demographics A and B applied 
to the same university U. Demographic parity is achieved only if the
percents of admitted students from the two demographics are equal, irrespective of whether one group is on average more qualified than the other.
\item \ib{Separation} type also considers that sensitive characteristics
are statistically independent of the model's prediction, but only when the
target/real value is provided. This is because this category acknowledges that
the sensitive feature may be correlated with the target variable. Metrics such
as equality of opportunity and equality of odds
\cite{hardtEqualityOpportunitySupervised2016a} fall under this subcategory.
Extending from our previous example, consider that students from demographics A
and B both applied to a science program in the university U. Students
from demography A had a better science curriculum in the high schools than the
students from demography B. Here, equality of odds would be satisfied if the
students who qualified, are equally likely to get admitted to the program, no
matter whether the students are from demography A or from demography B.
\item \ib{Sufficiency}: The conditional independence considered in this type of
fairness is actually opposite to Separation. It considers that given the model's
prediction, sensitive characteristics are statistically independent of the
target/real value. Metrics such as calibration within groups and predictive
parity \cite{chouldechovaFairPredictionDisparate2016} belong to this
subcategory. Consider that the example university U uses a model to predict the
acceptance probability of a student. If the model's precision rate is equal for
both demography A and demography B, then we can say the model satisfies the
predictive parity for demography.
\end{enumerate}

\textbf{Formal representation:} If we denote the set of all attributes as $X$, the real target value as $Y$, the predicted target value as $\hat{Y}$, and the set of sensitive attributes as $A$, then the criterion of the above metrics can be formulated as:
\begin{itemize}
\item Independence: \(\hat{Y} \bigCI A\)
\item Separation: \(\hat{Y} \bigCI A | Y\)
\item Sufficiency: \(Y \bigCI A | \hat{Y}\)
\end{itemize}
Here, $\bigCI$ denotes independent of and $|$ denotes conditional to.

\nd\bf{\ul{Group-Individual fairness metrics}} can
find bias in both group-based and individual cases. Different entropy indices,
such as coefficient of variation, entropy index, Theil index, etc. fall under
this detection metrics category \cite{speicherUnifiedApproachQuantifying2018}.



\nd\bf{\ul{Subgroup fairness metrics.}} Kearns et al. find that ensuring
group fairness does not guarantee fairness when there exist multiple sensitive
attributes \cite{kearnsPreventingFairnessGerrymandering2018}. In this case,
subgroup fairness metrics can be used to reduce the gap between individual and
statistical notions of fairness. Metrics such as rich subgroup fairness
\cite{kearnsPreventingFairnessGerrymandering2018} or differential subgroup
fairness \cite{fouldsIntersectionalDefinitionFairness2019} fall under this
category. Revisiting the university U example, consider that students from demography A
come from 2 different high schools (school A1, and school A2). Ensuring group
fairness for demography A does not guarantee fairness for the students from
both schools. There might be cases where most of the students who got admitted
are from one of the schools only. Subgroup fairness metrics can alleviate this
problem.

\nd\bf{\ul{Individual fairness metrics}} focus on the mistreatment
of individuals instead of focusing on the demography or group, they belong to.
Metrics such as Consistency~\cite{zemelLearningFairRepresentations2013} and
similarity metrics (e.g., Euclidean distance, Mahalanobis distance
\cite{demaesschalckMahalanobisDistance2000}) can be used to find biased decisions
made for any individual. Generally, the approaches to measuring individual
fairness can be based on both fairness through awareness
\cite{dworkFairnessAwareness2011} where the ML system is aware of the protected
attribute(s) and their probable impacts on the decision-making process, or
fairness through unawareness \cite{kusnerCounterfactualFairness2017} (aka
blindness) where it is tested that the ML system is not explicitly using the
protected attribute(s) to make any decision. In the example of the university U
admission, the admission scenario would satisfy individual fairness if two
applicants with identical profiles (grades, test scores, etc.) would be
considered equally for university admission, irrespective of the demography
they belong to.

\nd\bf{\ul{Utility fairness metrics}} do
not fall under any specific category, because they include some supporting or
utility APIs to aid the detection of bias in any given scenario. The number of
false positives and the number of false negatives are such supporting metrics. These
metrics can be used to develop custom fairness metrics when needed. For example,
a utility metric (e.g., accuracy), can be used to develop the equal accuracy
fairness metric, which is a statistical group fairness
metric that tests whether the prediction accuracy of both privileged and unprivileged groups are equal or not~\cite{castelnovoZooFairnessMetrics2021}.


\subsection{Bias Mitigation Approaches}\label{sec:background-bias-mitigation}
Bias mitigation approaches are primarily divided into three types:
\begin{inparaenum}[(1)]
\item Pre-processing, 
\item In-proceessing, and 
\item Post-processing.
\end{inparaenum} We discuss further classification of these approaches under each type below.

\nd\textbf{\ul{Pre-processing Techniques}} focus on the training datasets and can be divided into three types.
\begin{enumerate}[leftmargin=10pt]
\item Data transformation approach transforms the dataset into another representation, 
where the impact of the protected attributes (e.g., gender) gets minimized. 
The learning fair representations approach~\cite{zemelLearningFairRepresentations2013}) finds 
a latent representation of the data which encodes the original data well but obfuscates information about the sensitive attributes. 
\item Resampling uses sampling techniques, such as upsampling, and reweighting, to achieve fairness based on an attribute set. 
The reweighing algorithm ~\cite{kamiranDataPreprocessingTechniques2012a} weights the examples in each \textit{group, label} combination of protected attributes differently to ensure that samples from each group receive fair weights which in turn improves the fairness of the trained classification model.
\item Relabelling relabels the dataset so that group fairness can be achieved. For example, the \textit{Relabeller} approach \cite{bantilanThemisMLSource2017}  from the ThemisML library uses the proximity of the input data from the decision boundary of the input data space to change the label of the input data point.
\end{enumerate}

\nd\bf{\ul{In-processing Techniques}} work during the ML training phase and apply various regularizations and
constraints on the models to make their output fairer. We divide the techniques into five types:
\begin{enumerate}[leftmargin=10pt]
\item Fairness constraints: This approach applies fairness constraints to the model training layer so that the model does not learn a biased representation of the overall problem. For example, the \textit{meta fair classifier} proposed by Celis \textit{et al.} \cite{celisClassificationFairnessConstraints2020} uses fairness metrics as constraints to train an optimized classifier that results in a more fair classifier.
\item Multi-objective optimization algorithms formulate bias as a multi-objective optimization problem. The \textit{adversarial debiasing} algorithm~\cite{zhangMitigatingUnwantedBiases2018} proposed by Zhang \textit{et al.} develops a classifier that maximizes the overall prediction accuracy while reducing an adversary's ability to determine the protected attribute from the model's outcome. 
\item Regularization techniques put regularizing constraints while training an ML model. For example, Kamishima \textit{et al.} proposed \textit{prejudice remover} \cite{kamishimaFairnessawareClassifierPrejudice2012} which adds a discrimination-aware regularization term to the learning objective of an ML model. 
\item Neutralizing probability distribution algorithm tries to neutralize the probability distribution of both privileged and underprivileged groups so that there exists no difference in their representation. For example, the discrimination-free classification~\cite{caldersThreeNaiveBayes2010} approach neutralizes the probability distribution learned by a classifier to reduce the discrimination in decision-making. 
\item Worst case error minimization: Generally, a machine learning algorithm maximizes the overall accuracy by minimizing the overall error. This approach, however, minimizes the worst-case error instead of the overall error. The \textit{distributionally-robust optimization}~\cite{hashimotoFairnessDemographicsRepeated2018}, proposed by \textit{Hashimoto et al.}, is an example of this approach. 
\end{enumerate}

\nd\bf{\ul{Post-processing Techniques}} are applied to a model's outcome to
reduce bias in the predictions. This approach is useful when the training
dataset is unavailable, or access to a model's training phase is not
possible---which is common when using external APIs. We divide the post-processing algorithms into two different types.
\begin{enumerate}[leftmargin=10pt]
\item Probabilistic label change: This approach uses probability distribution to learn which of the labels should be changed to make a fair outcome. For example, with an equalized odds objective metric (discussed in \sec\ref{sec:rq-bias-detection-strategies}), the calibrated equalized odds~ \cite{pleissFairnessCalibration2017} technique optimizes over calibrated classifier score outputs to find probabilities with which to change the labels of the predicted output so that the classifier output becomes fairer.

\item Random flipping: This method randomly changes the labels to make the overall model outcome fairer. For example, the EthicML library implemented the DP flip approach~\cite{EthicMLDocumentationMitigation}, which tries to achieve perfect demographic parity by randomly flipping the label of a number of predictions. As demographic parity is a fairness metric that calculates group fairness, thus applying random flipping to achieve demographic parity also mitigates group fairness issues.  
\end{enumerate}

\section{Study Setup}\label{sec:setup}
\begin{figure}[t] 
\centering 
\hspace*{-.2cm}
\includegraphics[scale=.8]{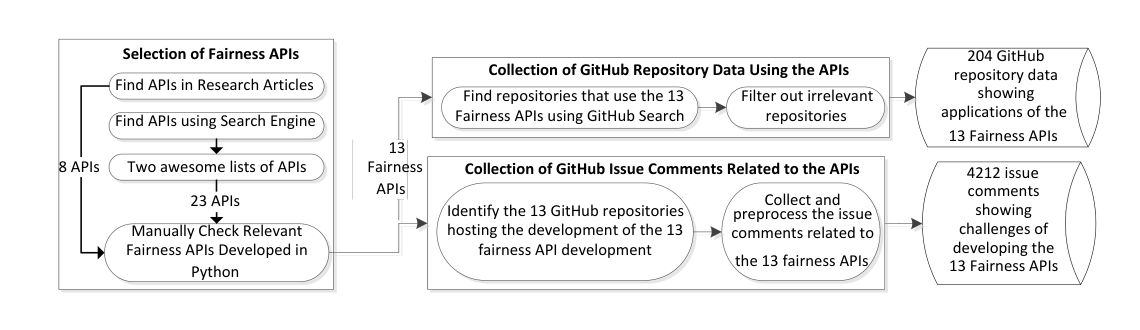}
\caption{ Major steps involved during the data collection of the study}
\label{fig:study_overview}
\end{figure}

In this paper, we answer four research questions (RQ):
\begin{enumerate}[label=\textbf{RQ\arabic{*}.}, leftmargin=30pt]
  \item What use cases the fairness APIs are applied for in the open-source Machine Learning Software Applications?
  \item How are the biases detected in these use-cases?
  \item How are the biases mitigated in these use-cases?
  \item What are the topics found in the issues that developers reported while using the APIs?
\end{enumerate}
To answer RQ1-RQ3, we utilize ML software repositories in GitHub 
and check their codebase to learn how 13 fairness API libraries are reused there. To answer RQ4, we use the issue discussions about the 13 fairness API libraries in 
their GitHub repositories.  

Hence, for our study, we collected data from GitHub. We selected GitHub because it is the most popular platform for open-source software development. As of
March 2025, more than 100 million developers have used GitHub and more than
420 million repositories are hosted on this platform.
A lot of ML repositories, including many of the most popular ML libraries (e.g.,
Sklearn, Tensorflow) are being hosted on GitHub.

The data collection process of our study involves three major steps as shown in \fig\ref{fig:study_overview} and summarized below.
\begin{enumerate}[leftmargin=30pt]
  \item \bf{Selection of Fairness API libraries (\sec\ref{sec:selection-apis}).} We identify 13 major fairness API libraries by systematically 
  consulting available public data containing the names of the fairness libraries (e.g., research articles, search engine outputs, and online awesome lists).
  \item \bf{Collection of GitHub Repository Data Using the API libraries (\sec\ref{dataCollection}).} We collect the usage of the 13 fairness API libraries from 204 GitHub ML software repositories.
  \item \bf{Collection of GitHub Issue Comments Related to the API libraries (\sec\ref{sec:collect-issue-comments}).} We collect 4212 issue discussions logged by the users/developers of the 13 API libraries from the GitHub repositories where the 13 API libraries are hosted. 
\end{enumerate} 

\subsection{Selection of the Fairness API libraries}\label{sec:selection-apis}

We need to find the sources of the available bias-related API libraries to understand how the bias detection and mitigation API libraries are used in the wild. 
Given there is no pre-defined approach to finding the list of available fairness API libraries, we followed a systematic approach to collect the names of fairness API libraries. The approach consists of three steps 
as shown in  \fig\ref{fig:study_overview}. First, we find a list of eight fairness API libraries by consulting research articles related to bias/fairness in ML. 
Second, we searched for fairness API libraries in the search engine Google, which pointed us to two publicly curated awesome lists for fairness APIs. The two awesome lists contain the name of a total of 23 fairness 
API libraries. The 23 API libraries also contain the eight API libraries that we found in the selected research articles. We discuss the search process for the research articles and Google below.

\nd\bf{\ul{Collect from research article:}} In this step, we searched for relevant research papers to collect available bias API libraries. 
We used a set of search phrases—``bias tool'', ``bias detection'', ``bias
mitigation'', and ``fairness tool''---in Google Scholar search to find the
relevant research articles. These phrases were selected after a discussion among
the first, second, and third authors. To be as relevant as possible and to
reduce false positives, we selected only the 10 most relevant research articles
from each of the search results. The first author went through the hit lists for
finding those articles. In the end, we found eight candidate libraries that were
potentially relevant to our study. \\
\nd\bf{\ul{Collect from Google search:}} In this step, we utilized the Google search engine to identify existing fairness API libraries. We used the same set of keywords used earlier to find relevant websites. We examined the first 10 results returned by Google search and listed all the API libraries we were able to collect.  The Google search also returned two popular and well-maintained lists of fairness-related API libraries \cite{CuratedListAwesome, CuratedListAwesomea}, which are commonly known as awesome lists. 
We included all the fairness-related libraries from these two lists for the following two reasons. \\
i) These two lists were initiated and maintained by researchers from two different research universities: Rice University and George Washington University, USA. \\
ii) Previous empirical studies also utilized awesome lists in their data collection process. For example, Gonzalez et al. \cite{gonzalezStateMLuniverse102020} used awesome lists to classify their machine learning repository dataset. \\
It is important to note that one of the awesome lists \cite{CuratedListAwesomea} was primarily developed for ML interpretability. However, we found that this awesome list contains libraries/toolkits related to fairness analysis, which made it relevant to our study. After completing this phase, we were able to collect 23 candidate libraries. Interestingly, all of our previous eight libraries (from the first phase) were contained within this new list.

\nd\bf{\ul{Filtering relevant API libraries:}}
Among the listed API libraries, we only considered the API libraries that provide bias detection or mitigation APIs, are open-source, and can be imported as Python packages.
In this study, we focus only on Python for several reasons.
\begin{inparaenum}[(i)] 
\item When we searched for fairness-related API libraries, we found that most of the popular fairness libraries and toolkits are written as Python APIs; 
\item Python is not only one of the most used programming languages for overall
software development \cite{StackOverflowDeveloper}, it is one of the de-facto
programming languages for ML and data-science
projects~\cite{gonzalezStateMLuniverse102020}.
\end{inparaenum}

After selecting the candidate list of fairness API libraries, we had to exclude some additional API libraries. 
For example, we excluded the library named \emph{scikit-fairness} because it stated in its 
GitHub source repository that it was renamed to the library named \emph{Fairlearn} which we already listed in our 
candidate list of selected API libraries. We also excluded two API libraries (named \emph{Responsibly} and \emph{Parity}). 
These two API library names are very common even outside of the scope of the fairness domain. As a result, there already exists Python 
libraries that use similar import statements as the import statements of the relevant fairness libraries (`responsibly' 
for the Responsibly library and `parity' for the Parity library). For this reason, including these libraries produces lots of 
false positives in our repository searching stage when we select candidate software repositories that use these fairness libraries. 
For example, 
the repository \href{https://github.com/rehive/aion-web3}{rehive/aion-web3}
 uses a blockchain-related library named parity, and this repository is not a
 fairness-related repository. For mitigating the issue of manually filtering out a lot of false positives we decided to exclude these two libraries.



Finally, 13 API libraries were selected for our study which are presented in Table~\ref{tab:libraries}. 
The API libraries are sorted by their GitHub star counts. 
For example, the most popular fairness API library in our dataset, in terms of GitHub stars, is AI Fairness 360~\cite{AIFairness3602018}, developed by IBM. 
The third column `Type' denotes the two types of fairness services offered by these API libraries, i.e., detection of bias and mitigation of the bias. 
The last column `Description' offers a brief description of the API libraries. 
We find that seven out of 13 API libraries offer both detection and mitigation services (e.g., AI Fairness 360, Fairlearn, etc.), while the rest four API libraries 
only offer detection services (e.g., Aequitas).  

\begin{table}[t]
    \centering
    \rowcolors{1}{}{lightgray!30}
    \caption{Description of the selected bias detection and mitigation API libraries (\#Stars are collected from GitHub). An API library can offer two types of fairness services: Detection (D) and Mitigation (M)}
    \label{tab:libraries}
    \begin{tabular}{lllp{9cm}}
    \toprule
    \textbf{API Library}   &  \textbf{\#Stars}  &  \textbf{Type}  & \textbf{Description} \\ \midrule
    AI Fairness 360 \cite{AIFairness3602018}        &  1679               &  D, M &  An extensible library developed by IBM that contains various  
    bias detection, mitigation APIs proposed by the research community. \\ 
    Fairlearn \cite{FairlearnPythonPackage2018}              &  1230            & D, M &  A community-driven project to assist data scientists  
    improve fairness of AI systems. \\ 
    Black Box Auditing \cite{BlackBoxAuditing2015}     &  121     & D, M  & Uses a black box approach to audit and repair disparate impact bias.  \\ 
    Aequitas \cite{BiasFairnessAudit2018}                &  464             & D  &   Bias audit toolkit developed by University of Chicago that can be used to  
    audit the predictions of machine learning-based risk assessment tools.\\ 
    Fairness Indicators  \cite{FairnessIndicatorsTensorflow2019}   &  248  & D  &  A TensorFlow toolkit that can be used in evaluating, improving, 
    and comparing models for fairness concerns. \\ 
    FairML \cite{adebayoFairMLAuditingBlackBox2016}                 &  325               & D  &  Toolbox for auditing predictive models by quantifying the relative  
    significance of the model's inputs. \\ 
    Fairness Comparison \cite{FairnessComparisonComparing2017}    &  134             & D, M  &  Facilitates the benchmarking of 
    fairness-aware ML algorithms. \\ 
    Themis ML \cite{bantilanThemisMLLibrary2017}              &  91           & D, M  & Implements fairness-aware machine learning algorithms, 
    built on top of Pandas and Sklearn. \\ 
    EthicML  \cite{thomasEthicMLFeaturefulFramework2019}               &  14              & D, M  &  For researcher for performing and assessing algorithmic fairness. \\ 
    GerryFair \cite{GerryFairAuditingLearning2018}              &  28            & D, M  &  Focused on auditing and learning for subgroup fairness. \\ 
    Transparent AI  \cite{laugaTransparentAIPythonLibrary2019}        &  6        & D  &  A Library that aims to check ethics in artificial intelligence models. \\

    Fat-forensics \cite{sokol2022fat-forensics}  & 75 & D & Algorithmic Fairness, Accountability and Transparency Toolbox. \\

    Fairlens \cite{FairLens2022} & 90 & D & A library for automatically discovering bias and measuring fairness in data. \\

    \bottomrule
\end{tabular}%
\end{table}

In general, we found that these 13 API libraries can have three kinds of software modules that focus on three different tasks:
\begin{enumerate}
\item \emph{Detection metrics} to provide the functionality to identify and measure bias in an ML system;
\item \emph{Mitigation algorithms} to provide ways to mitigate bias or to improve fairness in an ML system; and 
\item \emph{Common datasets} to provide popular sample datasets so that the developers can use them to test the implementation of their domain-specific bias detection or mitigation approaches.
\end{enumerate}


\subsection{Selection of GitHub Machine Learning Software Repositories Utilizing the Fairness API libraries}
\label{dataCollection}

We identify repositories in GitHub that use the 13 selected
fairness API libraries in two steps: i) candidate
repository selection, and ii) filtering out irrelevant projects with Abstract Syntax
Tree (AST) analysis. We discuss the steps below.


\subsubsection{Collecting GitHub Repositories that Use the Selected Libraries} 

Our goal is to analyze ML software systems that use the 13 fairness API libraries to understand how the APIs are used in the real world. We pick the ML software systems from the repositories hosted on GitHub. 
To prepare the list of repositories that use the selected API libraries, we utilized the GitHub code search API\footnote{https://docs.github.com/en/rest/reference/search\#search-code}.
Due to the imposed limitation by the GitHub Code Search API, to search for the repositories where our 13 fairness API libraries are used, we tried two different queries:
\begin{inparaenum}[(1)] 
\item Adding the Python keyword ``import'' before the base import of an API library.  For example, for Fairlearn API library: ``import fairlearn language:python''; 
\item Using only the API library's base imports. For example, for AI Fairness 360 API library: ``aif360 language:python''.
\end{inparaenum}  
The first query only searches for import lines that ensure we are considering the library imports. In GitHub code searches, this query also includes cases such as ``from fairlearn import some\_function'' as it just searches 
whether both of these terms exist within the same line. In total we found 1472 repositories from this search query. \\
On the other hand, the second query ensures that we are also considering non-import lines as well. As it is not clear how and whether GitHub code search indexes all the code files, we concluded that searching for repositories using just the library names might also provide additional repositories to work with. In total we found 1526 repositories utilizing this search query which indeed returned a higher number of repositories than what the previous query returned. We conducted both of these queries on March 20, 2025. \\
In total, we obtained 1885 unique GitHub repositories by combining the repositories found using both of these approaches. In our final filtered dataset, we found 7 repositories (e.g., \href{https://github.com/iamgroot42/distribution_inference}{iamgroot42/distribution\_inference}, \href{https://github.com/wajrcs-dk/Fairness-aware-Machine-Learning}{wajrcs-dk/Fairness-aware-Machine-Learning}, etc.) that would have been missed if we had only used the import-based search query. This increases our confidence in using two different search queries.  Overall, these queries ensure that we will get as many relevant repositories as possible from the GitHub Code Search API. After finalizing our repository set, we then collected the source files of all of these repositories for further analysis.


\subsubsection{Identifying irrelevant repositories} 
We found that some of the
collected repositories can be irrelevant because:
\begin{inparaenum} \item The detected import lines are comments only; or
\item There exist valid library imports, but the API libraries were never actually used;
or \item The API is not a fairness API, even though the API name
matches with one of our selected libraries.
\end{inparaenum}
To resolve the first two issues, we follow an AST analysis-based approach which works in three steps.
\begin{itemize}[leftmargin=10pt]
\item \nd\bf{Search for the Python files:} As we focus only on Python
repositories, we searched for Python files within the repositories using the
\textit{\textbf{find}} utility of the Linux operating system. This is to
filter out irrelevant files, such as \texttt{txt}, and \texttt{json}.
\item \nd\bf{Search for the import lines:} We searched for all the import
statements within these Python files. We only kept the files that contain the
imports that match the base import modules of the targeted libraries for
subsequent analysis. We used the \textit{\textbf{grep}} utility to search for
these import statements within the source files.
\item \nd\bf{Analyze using the Abstract Syntax Trees (AST):} Then we analyzed
the detected files from the previous step by building the ASTs of their sources.

\item \nd\bf{Exclusion of private repositories:} To ensure the generalizability and relevance of our findings, we identified and excluded private repositories from the list. We considered a repository as private if it had only one contributor, and primarily focused on non-generic, real-world applications.
\end{itemize}
After building the AST of these Python files, we recursively searched in these
trees where the API imports were being used, and  whether any other module
was imported using the attribute calling mechanism, and how many times all these imported modules have been called inside these trees.
The AST analysis not only helped in removing noises but also helped us to find
out the exact detection and mitigation approaches that were used by the
repositories. 

The third type of noise --- identical names, but not a fairness library --- was
resolved by manual analysis. After this manual verification, we were left with 204
repositories that indeed used valid APIs from our selected list of API libraries.
This manual analysis process took $\sim$10 hours. All the analysis and close manual revision make the selection set free from false positives.

\subsection{Collection of Issue Comments Related to the Fairness API libraries}\label{sec:collect-issue-comments}



We analyzed the issues logged against the 13 fairness API libraries to understand the challenges developers face while using and/or developing the API libraries. 
An issue is a feature of the GitHub platform where the users and developers of a repository can report different types of issues (e.g., bug reports, and queries) regarding the API usage.  All the code repositories of the 13 selected fairness libraries are hosted on GitHub. The issue report contains the issue title, issue body, and all the comments that occurred within the particular issue. Using the GitHub API, we collected all the  989 issues (title, and body) and 3223 issue comments found within these 572 issue threads from 11 of the libraries. At the end, we produced a dataset of 4212 issue discussions.
We did not find any issues reported in GitHub for \emph{GerryFair}, and \emph{Black Box Auditing} APIs.

\section{Application of the Fairness APIs in the Machine Learning Systems (RQ1)}\label{sec:lib_usecases}
In this section, we present details of the application contexts and scenarios in
the studied ML software applications (aka MLSAs), where the fairness APIs are used. We answer the
following research question:

\nd\bf{RQ1. What use cases the fairness APIs are applied for in Machine Learning Software Applications?}

We divide this research question into the following sub-RQs as follows.
\begin{enumerate}[label=\textbf{RQ1.\arabic{*}.}, leftmargin=40pt]
  \item \bf{Pipeline.} How are the fairness APIs reused within an ML pipeline in the studied systems?
  \item \bf{Context.} What are the reasons behind the reuse of fairness APIs in the studied systems?   
  \item \bf{Scenario.} What application scenarios are tested for bias in the MLSAs using the fairness APIs?   
\end{enumerate} 
We start with explaining the ML pipelines in the studied systems where a fairness API is reused. 
We then check how the fairness APIs are reused within the implemented pipeline. 
Previous research has shown concerns about the possibility of bias in different types of ML models:
models for recruiting systems \cite{AmazonScrapsSecret}, for loan and credit approval systems \cite{chenFairLendingNeeds2018}, and for health care systems \cite{vokingerMitigatingBiasMachine2021}. 
In this research question, we investigate this issue in more detail by finding the distribution of use-cases, domains, and activities that deal with bias.

\subsection{RQ1.1 Fairness ML Pipeline}
\subsubsection{Motivation} 
In ML system development, ML developers usually perform certain common tasks
such as data preprocessing, feature engineering, model training, model
evaluation, etc. in almost all ML projects. Moreover, the ML projects perform
these tasks in step by step manner and we can combine these steps together into
a pipeline format. We call such a pipeline that contains stages of ML
development, ML pipeline, and the steps that build up this pipeline ML pipeline
stages.
On the other hand, fairness APIs have been developed to detect and mitigate
fairness-related issues with AI/ML systems or datasets. As there exist many
different stages in the ML system development pipeline, these fairness APIs can
also be applied in many different stages. Studying the pipeline stages where the
ML developers used the fairness APIs most frequently will provide us with
insights that other API users of these API libraries can leverage to explore
their options for integrating such APIs.

%

\subsubsection{Approach} 

To study the usage of fairness APIs with the machine learning pipeline, in the selected list of 204 repositories, we identified three possible stages where fairness APIs have been applied. The trends of usage fairness stages are: Data labeling, before model training, and within the model evolution level. The most popular level of use is the "Data labeling" level. 67\% of the repositories applied fairness in detecting or optimizing bias at the data level. 63\%  of the repositories applied fairness APIs before model training, and 43\%  of the repositories applied fairness APIs after the model evaluation stage. It is to mentioning that most of the repositories applied fairness in multiple stages. To design the study, we identified the known keywords (i.e., function calls, search keys, etc ) for each of the stages and searched within the target repositories. We considered the source code, the code documentation,
 and the repository description to locate occurrences of the identifiers.

\begin{figure}[!thb]
\centering
\includegraphics[scale=.75]{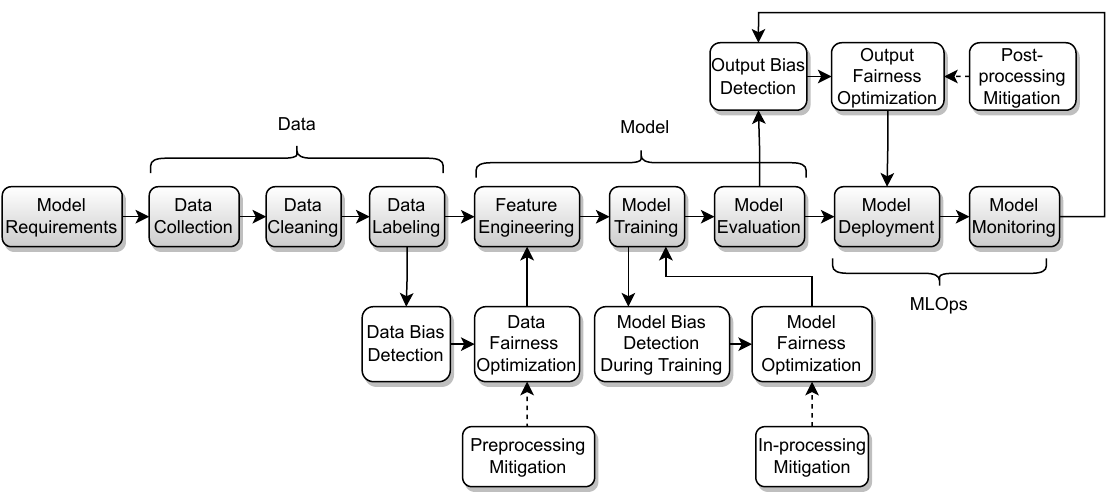}
\caption{The nine stages of the machine learning workflow (as presented by Amershi et al.~\cite{amershiSoftwareEngineeringMachine2019}).}
\label{fig:ml_pipeline}
\end{figure}

\subsubsection{Results}
As we visualized the ML pipeline used in our studied ML software systems, we 
found that the pipelines offer customization and/or extension over a traditional ML pipeline.
For a traditional pipeline, we follow the ML pipeline architecture presented by Amershi et
al.~\cite{amershiSoftwareEngineeringMachine2019} as the baseline architecture.
\fig\ref{fig:ml_pipeline} provides a flowchart of the ML pipeline stages as presented by Amershi et al.~\cite{amershiSoftwareEngineeringMachine2019}. Here we see that at the very beginning, the requirements of the model to be developed
are collected, and then relevant data collection, data cleaning, and data labeling are performed. After that, the core model development begins with feature engineering, followed by model training and model evaluation. Finally, the model is deployed and the model performance is monitored.
Along with the pipeline stages presented by Amershi et
al.~\cite{amershiSoftwareEngineeringMachine2019} (shown in darker boxes), in
\fig\ref{fig:ml_pipeline}, we also show the stages where our bias detection and
mitigation approaches can be incorporated into the existing ML pipeline.

Our study recognized that fairness API deals with data optimization, feature optimization, and model optimization. Based on the findings we devised that, bias detection and mitigation APIs can be incorporated into the existing pipeline stages in the following ways:

\begin{enumerate}
    \item after the data labeling stage, for performing data fairness optimization;
    \item within the model training stage, for performing model fairness optimization;
    \item finally, after the model evaluation stage, for performing the ML model's output fairness optimization.
\end{enumerate}

\begin{table*}[t]
    \centering
    \caption{ML pipeline stages within which the Fairness APIs have been reused.}
    \rowcolors{1}{}{lightgray!30}
    \label{tab:ml_pipeline}
    \scalebox{0.9}{
    \begin{tabular}{lcp{11cm}}
        \toprule
        \textbf{ML Pipeline Stage}  & \textbf{\% of Repos} & \textbf{Example Repos} \\ \midrule
        After Data Labeling  &   66\% &  Fairness Indicator (\href{https://github.com/henriquepeixoto/Data-science-cool-stuff}{henriquepeixoto/Data-science-cool-stuff}), AI Fairness 360
 (\href{https://github.com/pnb/fairfs}{pnb/Fairfs}) \\
    
        After Feature Extractions   &   63\%  &  Grayfair(\href{https://github.com/lacava/fair_gp}{lacava/fair\_gp}), AI Fairness 360 (\href{https://github.com/JohannesJacob/fairCreditScoring}{JohannesJacob/fairCreditScoring})  \\
        
        After Model Evaluation    & 43\% & AI Fairness 360 (\href{https://github.com/kserve/kserve}{kserve/kserve}), Fairlearn (\href{https://github.com/pycaret/pycaret}{pycaret/pycaret}), \\
        \bottomrule
    \end{tabular}
    }
\end{table*}

We present the summary of the findings of the ML pipeline stage study in
\tbl\ref{tab:ml_pipeline}. The findings suggest that, fairness APIs are applied mostly in multiple stages of ML pipeline and Data level optimization is the most common. We describe these fairness optimization stages in detail along with relevant code examples found within our studied repositories.

\nd\bf{1) Model input data fairness optimization:} Within these stages, at first, fairness detection APIs are applied to detect fairness issues within the dataset itself. Then pre-processing bias mitigation algorithms/APIs are applied to mitigate any detected fairness issues and to improve the fairness of the relevant dataset. Finally, the fair dataset is used within the rest of the pipeline stages. 
The repository 
\href{https://github.com/henriquepeixoto/Data-science-cool-stuff}{henriquepeixoto/Data-science-cool-stuff}
 has used APIs that fall under the data fairness optimization stage.
This repository has used the utility metric (i.e.,
\textit{util.convert\_comments\_data}) from the Fairness Indicator library to
pre-process the used dataset (as presented in
Listing~\ref{lst:pipeline_data_prep}). However, the API that this repository
used is a utility function, and it does not have any fairness-related impact.

\begin{lstlisting}[caption={Applying the fairness APIs in the data 
preprocessing stage in repository \href{https://github.com/henriquepeixoto/Data-science-cool-stuff}{henriquepeixoto/Data-science-cool-stuff}},label={lst:pipeline_data_prep},basicstyle=\footnotesize]
.....
.....
from fairness_indicators.examples import util
.....
.....
train_tf_file = <@\textcolor{black}{\colorbox{yellow}{util.convert\_comments\_data}}@>(train_tf_file)
validate_tf_file = <@\textcolor{black}{\colorbox{yellow}{util.convert\_comments\_data}}@>(validate_tf_file)
.....
.....
\end{lstlisting}

On the other hand, we also explored additional repositories to find better code examples of the usage of the APIs within these bias detection and mitigation stages. For this exploration, we used the insights gained from the detected bias detection and mitigation APIs (discussed in detail in \sec\ref{sec:rq-detection-mitigation}) to find these relevant repositories as these detection and mitigation APIs are related to specific ML stages. For example, we find the following example (shown in Listing~\ref{lst:pipeline_data_prepro}) within the \href{https://github.com/danladishitu/CE888-7-SP}{danladishitu/CE888-7-SP} repository where the author used the pre-processing mitigation approach, \textit{Reweighing} from the AI Fairness 360 library, to mitigate bias within the input dataset.

\begin{lstlisting}[caption={Applying the fairness APIs to optimize data fairness in repository \href{https://github.com/danladishitu/CE888-7-SP}{danladishitu/CE888-7-SP}},label={lst:pipeline_data_prepro},basicstyle=\footnotesize]
......
......
from aif360.datasets import BinaryLabelDataset 
from aif360.metrics import ClassificationMetric
from aif360.explainers import MetricTextExplainer  
from aif360.algorithms.preprocessing import Reweighing
......
......
valid_metric = ClassificationMetric(bld, bld_preds, 
            unprivileged_groups=unprivileged_groups,
            privileged_groups=privileged_groups)

balanced_accuracy.append(0.5 * (valid_metric.true_positive_rate()
            + valid_metric.true_negative_rate()))
avg_odd_diff.append(valid_metric.average_odds_difference())
disp_impact.append(np.abs(valid_metric.disparate_impact() - 0.5))
......
......
rw = <@\textcolor{black}{\colorbox{yellow}{Reweighing}}@>(unprivileged_groups=unprivileged_groups,
                privileged_groups=privileged_groups)
train_pp_bld_f = rw.fit_transform(train_pp_bld)

# Create the metric object
metric_train_bld = BinaryLabelDatasetMetric(train_pp_bld_f,
                                           unprivileged_groups=unprivileged_groups,
                                            privileged_groups=privileged_groups)
......
......
\end{lstlisting}

\nd\bf{2) Model design fairness optimization:} Within this stage, the fairness of the trained ML model itself is targeted, and in-processing mitigation approaches are applied to optimize the fairness aspects of the developed ML model. 
In Listing~\ref{lst:pipeline_model_inpro} we show how the repository
\href{https://github.com/JohannesJacob/fairCreditScoring}{JohannesJacob/fairCreditScoring}
 has applied an  in-processing mitigation algorithm,
\textit{PrejudiceRemover} from the AI Fairness 360 library, to optimize the fairness of the trained ML model.

\begin{lstlisting}[caption={Applying the fairness APIs to optimize model fairness in repository \href{https://github.com/JohannesJacob/fairCreditScoring}{JohannesJacob/fairCreditScoring}},label={lst:pipeline_model_inpro},basicstyle=\footnotesize]
......
......
from aif360.algorithms.inprocessing import PrejudiceRemover
......
......
all_eta = [1, 5, 15, 30, 50, 70, 100, 150]

for eta in all_eta:
    print("Eta: %.2f" % eta)
    colname = "eta_" + str(eta)

    debiased_model = <@\textcolor{black}{\colorbox{yellow}{PrejudiceRemover}}@>(eta=eta, sensitive_attr=protected, class_attr = "TARGET")
    debiased_model.fit(dataset_orig_train)
    
    dataset_debiasing_valid = debiased_model.predict(dataset_orig_valid)
    dataset_debiasing_test = debiased_model.predict(dataset_orig_test)
......
......
\end{lstlisting}

\nd\bf{3) Model output fairness optimization:} Fairness detection and optimization stages can also be applied within the ML model's generated output to eradicate fairness issues within the generated model output. The post-processing bias mitigation algorithms are applied within this stage to improve the fairness of the generated ML output. 
For example, in Listing~\ref{lst:pipeline_output_detect} we show how the
repository
\href{https://github.com/predictive-analytics-lab/nifr}{predictive-analytics-lab/nifr} utilized the EthicML library to
evaluate the fairness of the trained model's output. Here we see that the
repository has used various fairness metrics such as \textit{True Positive Rate (TPR)},
\textit{Theil Index}, etc., to calculate the fairness aspects of the generated output.

\begin{lstlisting}[caption={Applying the fairness APIs to evaluate output fairness in repository \href{https://github.com/predictive-analytics-lab/nifr}{predictive-analytics-lab/nifr}},label={lst:pipeline_output_detect},basicstyle=\footnotesize]
......
......
from ethicml.algorithms.inprocess import LR, SVM, Agarwal, Kamiran, Majority
from ethicml.evaluators.evaluate_models import run_metrics
from ethicml.metrics import CV, NMI, PPV, TNR, TPR, Accuracy, ProbPos, Theil
from ethicml.preprocessing.train_test_split import train_test_split
......
......
for mix_fact in [k / 100 for k in range(0, 105, 5)]:
   args = _Namespace()
   args.task_mixing_factor = mix_fact
   _, task, task_train = load_adult_data_tuples(args)
   try:
       preds = clf.run(task_train, task)
   except:
       print(f"{clf.name} failed on mix: {mix_fact}")
       continue
   metrics = <@\textcolor{black}{\colorbox{yellow}{run\_metrics}}@>(preds, task, metrics=[Accuracy()], per_sens_metrics=[])
   print(f"{clf.name} Accuracy: {metrics['Accuracy']}")
   res_dict = run_metrics(
       preds,
       task,
       metrics=[<@\textcolor{black}{\colorbox{yellow}{Accuracy(), Theil(), NMI(), TPR(), TNR(), PPV()}}@>],
       per_sens_metrics=[<@\textcolor{black}{\colorbox{yellow}{Theil(), ProbPos(), TPR(), TNR(), NMI(), PPV()()}}@>],
   )
   res_dict["mix_factor"] = mix_fact
......
......
\end{lstlisting}

In Listing~\ref{lst:pipeline_output_postpro}, we show how the
\href{https://github.com/JohannesJacob/fairCreditScoring}{JohannesJacob/fairCreditScoring}
repository applied post-processing mitigation algorithms,
\textit{RejectOptionClassification} and \textit{EqOddsPostprocessing}, to
optimize the fairness of the generated output.

\begin{lstlisting}[caption={Applying the fairness APIs to optimize the output fairness in repository \href{https://github.com/JohannesJacob/fairCreditScoring}{JohannesJacob/fairCreditScoring}},label={lst:pipeline_output_postpro},basicstyle=\footnotesize]
......
......
from aif360.metrics import BinaryLabelDatasetMetric
from aif360.algorithms.postprocessing.reject_option_classification\
        import RejectOptionClassification
from aif360.algorithms.postprocessing.eq_odds_postprocessing\
        import EqOddsPostprocessing
......
......
# Metric used (should be one of allowed_metrics)
metric_name = "Statistical parity difference"
......
......
model_names = ['glm', "svmLinear", "rf", "xgbTree", "nnet"]
......
......
for m in model_names:
    ......
    ......
    # Reject Option Classification
    ROC = <@\textcolor{black}{\colorbox{yellow}{RejectOptionClassification}}@>(unprivileged_groups=unprivileged_groups, 
                                     privileged_groups=privileged_groups, 
                                     low_class_thresh=0.01, high_class_thresh=0.99,
                                      num_class_thresh=100, num_ROC_margin=50,
                                      metric_name=metric_name,
                                      metric_ub=metric_ub, metric_lb=metric_lb)
    ROC = ROC.fit(dataset_orig_valid, dataset_orig_valid_pred)
      
    # ROC_test results
    dataset_transf_test_pred = ROC.predict(dataset_orig_test_pred)
    ......
    ......
    # Equality of Odds
    EOP = <@\textcolor{black}{\colorbox{yellow}{EqOddsPostprocessing}}@>(unprivileged_groups=unprivileged_groups, 
                               privileged_groups=privileged_groups,
                               loss_name="Balanced accuracy groupwise")
    EOP = EOP.fit(dataset_orig_valid, dataset_orig_valid_pred)
    
    # EOP_test results
    dataset_transf_test_pred = EOP.predict(dataset_orig_test_pred)
......
......
\end{lstlisting}

In summary, we see that fairness APIs have been used in various stages of the ML pipeline ranging from optimizing the fairness of the used training dataset to improve the fairness of the generated ML model output. As such the usage of these fairness API libraries covers a wide ground within the ML model development life cycle.

\subsection{RQ 1.2 Fairness Application Contexts}\label{sec:app_contexts}
\subsubsection{Motivation}
We have already seen that there exist various fairness APIs that can be used in
various ways to aid in detecting and mitigating fairness-related issues.
However, we have not seen any study that discusses the purpose of using these
fairness APIs within real-world repositories. In this subsection, we discuss the
primary reasons for which the fairness APIs had been used within our selected set of 204 repositories.


\subsubsection{Approach}
To discover the reasons behind the usage of fairness APIs within these
repositories, we begin at the ground level by finding the use cases of the fairness API libraries. 
To find out the use cases of the fairness API libraries, we label each of the selected 204
software repositories based on their intended actions. For this labeling, we
applied an open card sorting approach \cite{williamCardSorting2013}. The first and third
authors (a graduate student and a postdoc) worked together during the
labeling phase. These two authors synchronously communicated using Skype and
Zoom meetings. They have discussed the name, description, and README.md file of
each repository from GitHub using GitHub APIs.
In a few cases, specifically, when the projects are developed for tutoring
purposes, the intended use-cases were not clear from the metadata.
The two authors went through the source files of the repositories to find out
the intended use-cases.
More specifically, they have used project reports or documentation files, the
used fairness datasets, and APIs, the comments found within the driver file, and
the source code fragments where the bias-related APIs were used. 

For example, the GitHub repo \href{https://github.com/SotArchontas/AutoML}{SotArchontas/AutoML} 
has been labeled as `ML Development Framework' use-case. The purpose of the auto ML framework is to 
develop a machine-learning pipeline using different automatic techniques. The 
repository \href{https://github.com/SotArchontas/AutoML}{SotArchontas/AutoML} has added components to apply 
fairness constraints while developing different ML models. However, to test its generated pipeline using fairness 
constraints it used multiple fairness benchmark datasets from different domains (such as business, and health). 
As such, we see datasets from different domains even though the primary target of the repository is to develop a 
machine-learning model using an automated approach. This is why we labeled this repository under the `ML Development Framework' use-case.

Similarly, another GitHub repo \href{https://github.com/dssg/hiv-retention-public}{dssg/hiv-retention-public} uses the Aequitas fairness 
API to detect bias in customer retention in clinical trials. We thus label the use case as `Patient Retention'.
Overall, the labeling of the uses took the two authors $\sim$25 hours in total. 
The second author was consulted to finalize labeling in various stages, including the generation of a taxonomy of the identified use cases.


Finally, we reviewed these use cases and derived the primary motivations behind the usage of these fairness APIs within this repository set.

\subsubsection{Results}
In our selected set of relevant repositories, we see two primary reasons for using fairness APIs within these repositories. We see that the APIs have been used-
\begin{inparaenum}[(1)]
    \item for generic learning and exploration purpose;
    \item for solving real-world problems.
\end{inparaenum} 
Here we discuss these purposes in greater detail.
\\
\indent\bf{1) Generic learning and exploration purpose:} Majority (70.59\% of all repositories) of the repositories within our selected set of 204 repositories have used the fairness APIs for learning and exploration purposes. The primary focus within these repositories was to create tutorials to support the learning of a fairness API or the implementation of a toy service using these bias APIs so that the repository creators could explore the functionalities of these fairness libraries.

\begin{figure}[!thb]
\centering
\includegraphics[scale=.8]{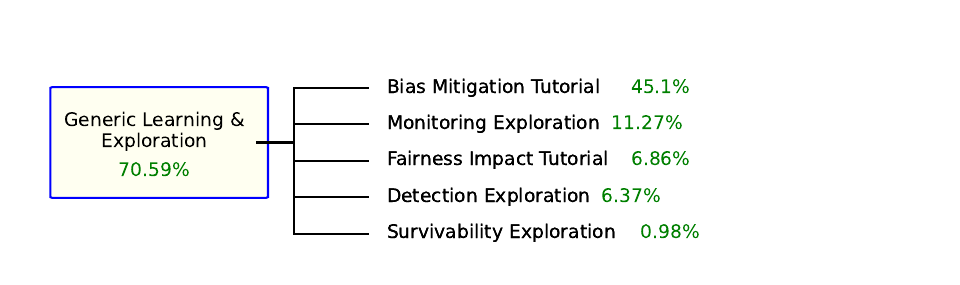}
\caption{ Taxonomy of the bias-related use-cases that served the generic learning and exploration purpose within the open source ML software systems that apply bias detection and/or mitigation APIs.}
\label{fig:usecase_categ_generic}
\end{figure}

For example, within our selected repository set, there are projects that explore different bias detection and mitigation 
algorithms (\href{https://github.com/andrijaster/FAIR}{andrijaster/FAIR}). These approaches are not focused on any specific domain and can be applied in various areas. The primary objective found in such repositories is to provide learning/educational materials.
As such, the generic purpose use cases are mostly implemented in notebooks, unlike the non-generic use cases (see below). 
\fig\ref{fig:usecase_categ_generic} summarizes the taxonomy of the 5 unique use cases that can be found on repositories that focus on using the APIs for learning and exploration purposes,
including repositories with educational materials and learning
activities. 
 The use cases are described below:
\begin{itemize}

    \item \textbf{Bias Mitigation Tutorial}: This category includes repositories that guide mitigating bias in machine learning models. Approximately 45.1\% of the repositories fall under this classification. For instance, the repository \href{https://github.com/andrijaster/FAIR}{andrijaster/FAIR} offers an example of how to train models to learn unbiased representations using backpropagation. Unlike general educational tutorials, which often introduce various libraries and present academic comparisons, bias mitigation tutorials are specifically focused on tools and APIs developed to detect and reduce bias in machine learning systems.
    \item \textbf{Monitoring Explorations}: This category includes repositories focused on monitoring the fairness of machine learning APIs. An example is the repository \href{https://github.com/aideenf/machnamh}{aideenf/machnamh}, which facilitates reflective analysis of machine learning models to assess and track fairness over time. 

   \item \textbf{Fairness Impact Tutorials}: This category comprises repositories that focus on analyzing the impact of fairness-related APIs on machine learning systems. For example, the repository \href{https://github.com/pernillej/Cost-of-Fairness}{pernillej/Cost-of-Fairness} examines the trade-offs and cost implications associated with incorporating fairness into AI-based decision-making models. 

    \item 
    \textbf{Detection Exploration}: The repositories engaged in bias detection-related processes were aligned to this category. For example, \href{https://github.com/Call-for-Code-for-Racial-Justice/bias-detection-engine}{Call-for-Code-for-Racial-Justice/bias-detection-engine} prepared a bias and disparity detection engine to isolate disparity in Federal sentencing outcomes of Black vs White defendants of the United States.

    \item 
    \textbf{Survivability Exploration}: This category includes repositories that analyze specific populations' survivability using data-driven approaches. These projects typically assess how fairness-related factors influence survival outcomes. For example, \href{https://github.com/mbunse/mlcomops}{mbunse/mlcomops} mitigated the bias in survivability prediction.
    
\end{itemize}

\bf{2) Solving a real-world problem:} The other reason why fairness APIs have been used within these repositories is indeed to address the fairness aspects of various real-world ML applications and datasets. Various bias detection approaches have been used in these repositories to detect fairness issues in the trained ML models and datasets. Along with that, various bias mitigation methods and algorithms have been utilized to reduce the impact of bias and improve fairness within ML applications. As a result, these repositories could provide us with more insights into the application contexts of our studied fairness API libraries. As a result, we discuss such use cases in detail in \sec\ref{sec:app_scenarios}, explaining with detailed examples. 
The

\subsection{RQ1.3 Fairness Application Scenarios}\label{sec:app_scenarios}

\subsubsection{Motivation}
In the previous subsection (\sec\ref{sec:app_contexts}), we discuss the two
primary reasons behind the reuse of the fairness APIs within our selected
repository set. In this subsection, we dig deeper into the second reason for using the fairness APIs, serving the non-generic real-world use cases
that address the fairness aspects of various ML applications and datasets. 

\subsubsection{Approach}
In \sec\ref{sec:app_contexts}, we mention that we started our application context analysis phase by first finding out the use cases of our selected set of repositories. Then, we classified the reason behind the usage of the fairness APIs within these repositories into two categories- generic learning purpose and real-world applications. Within these non-generic real-world application repositories, we observe that these use cases can also be classified into certain activity types. These activity types represent how the ML systems developed within these repositories are addressing these use cases. For example, some ML systems focus on prediction tasks/activities to address certain use cases, other systems focus on analysis activities to achieve their respective use cases. For example, the `Patient Retention' use case is about the prediction of patients that should or should not be retained within medical institutions. We see that fairness APIs have been used within this specific use case to identify and/or remove bias within the patient retention calculation process. Therefore, we label the activity type for this use case as `Prediction'.  
Similarly, another use case `ML Development Framework' belongs to the activity type `Operation', given it is about the operationalization of an ML model. The fairness APIs have been used within this use case in incorporating fairness constraints into the model development pipeline.
In total, the third author worked $\sim$10 hours with the first two authors for this categorization.

\subsubsection{Results}
\begin{figure}[!thb]
\centering
\includegraphics[scale=.9 ]{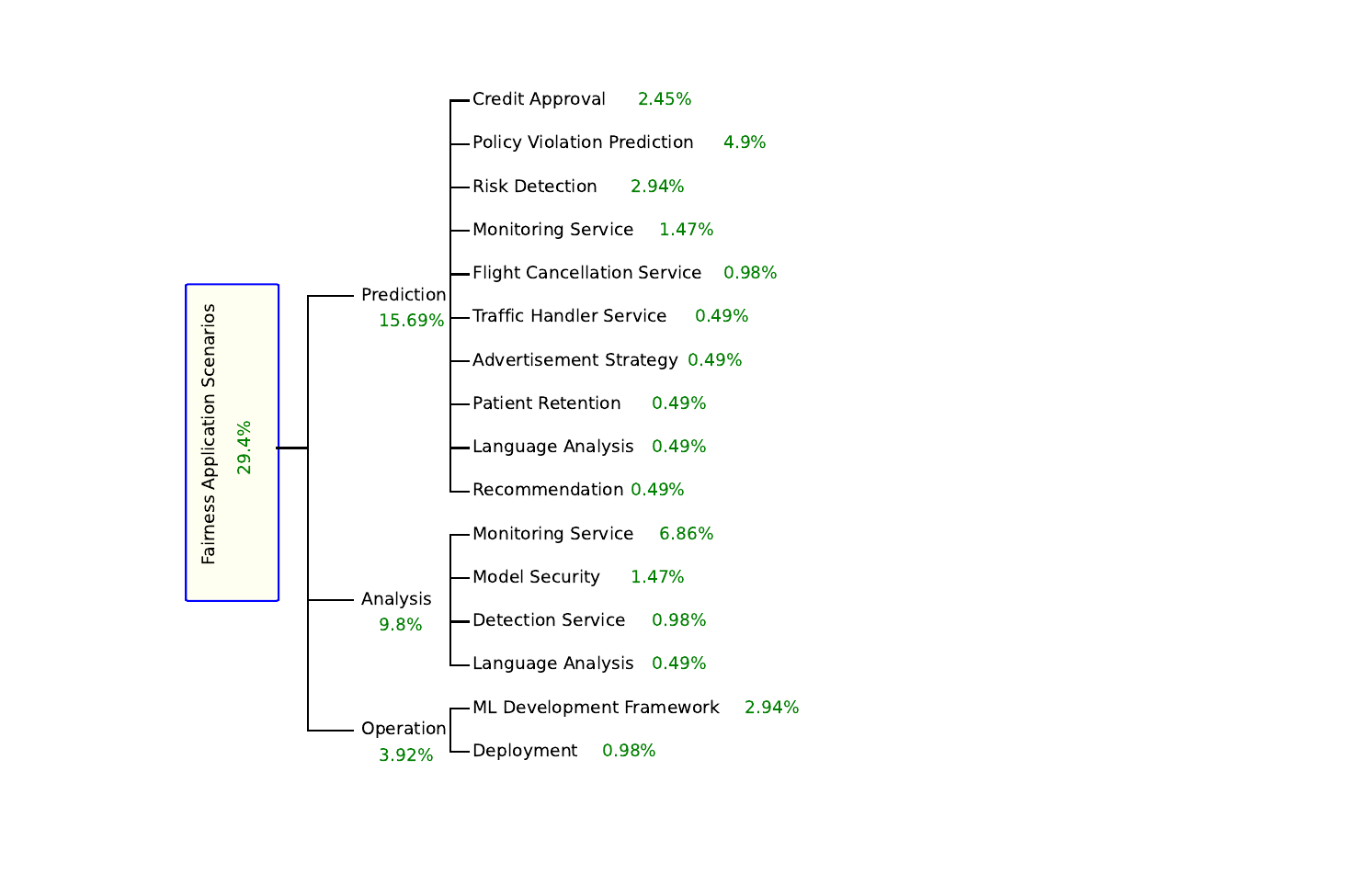}
\caption{Taxonomy of the bias-related use-cases that served the non-generic application purpose within the open source ML software systems that apply bias detection and/or mitigation APIs.}
\label{fig:usecase_categ_nongeneric}
\end{figure}

Within our selected repository set, 29.4\% of the repositories contain ML systems that served various non-generic real-world ML applications. \fig\ref{fig:usecase_categ_nongeneric} summarizes the taxonomy of the use-cases of the repositories that focused on these non-generic applications. These software repositories 
utilized the fairness APIs for solving fairness issues in addressing genuine real-world tasks.
Within this filtered repository set, we find that the use cases `Monitoring Service' and `Credit Approval' are the most frequent use cases. We discuss below the non-generic use cases by explaining the activities and their served use cases in detail.

\nd\bf{$\bullet$ The three primary activity types found within the non-generic repositories are \begin{inparaenum}[(1)]
\item Prediction, 
\item Analysis, and 
\item Operation.
\end{inparaenum}}

\begin{inparaenum}[(1)]
\item The \emph{Prediction} category contains projects (15.69\%) that are focused on prediction tasks.
For example, the project \href{https://github.com/satejsoman/chicago-business-viability}{satejsoman/chicago-business-viability} used bias detection APIs to identify bias in a business failure prediction task, and therefore enlisted under this category. A total of 10 different use cases belong to this category. These use cases cover various topics ranging from financial prediction to performing prediction in the health domain.  

\item The \emph{Analysis} category contains projects (9.8\%) that analyze data and ML models. 
For example, the project \href{https://github.com/johntiger1/multimodal_fairness}{johntiger1/multimodal\_fairness} aims at analyzing the impact of text-specific and blackbox fairness algorithms in the domain of Multimodal Clinical NLP.

\item The \emph{Operation} category is the least prevalent and contains projects (3.92\%) that provide support during ML system operation. These activities were mainly focused on the development of ML models (i.e., \href{https://github.com/SotArchontas/AutoML}{SotArchontas/AutoML}), and the deployment (i.e., \href{https://github.com/IBM/Trusted-ML-Pipelines}{IBM/Trusted-ML-Pipelines}) of the developed models. 
\end{inparaenum}

In total, we find 17 distinct use cases within the selected repository set. Within these use cases some use cases, e.g., `Monitoring' and `Detection' exist across both generic learning purposes and non-generic repositories. For example, the `Monitoring' use-case can be found as `Monitoring Exploration' within the learning-purpose repositories and as `Monitoring Service' within the non-generic repositories. Out of these 17 use cases, 14 use cases belong to non-generic real-world applications.
Our replication package describes all of the use-cases that we show in \fig\ref{fig:usecase_categ_generic} and \ref{fig:usecase_categ_nongeneric}. 
To provide insights about our findings, we discuss below several selected non-generic use-cases, covering all three types of activities (i.e., Prediction, Analysis, and Operation).

\begin{inparaenum}[(1)]
\item \emph{Monitoring Service (Analysis)}: This use-case is related to projects
that focus on the analysis of System/API monitoring services. For example, for
specific datasets and prediction problems, the application project,
\href{https://github.com/nirdizati-research/predict-python}{nirdizati-research/predict-python},
provides support to experienced users in finding the most suitable
predictive models.

\item \emph{ML Development framework} (Operation) are projects related to developing a
pipeline for automated model development, checking bias while using AutoML tools
to develop models, neural network training, and certification. The project
\href{https://github.com/SotArchontas/AutoML}{SotArchontas/AutoML} falls into
this category because it is a project which integrates auditing of fairness
within the pipeline of AutoML-based low-code model development strategy.


\item \emph{Patient Retention (Prediction)} are projects that predict
patient's retention in clinical care (e.g.,
\href{https://github.com/dssg/hiv-retention-public}{dssg/hiv-retention-public}).

\item \emph{Risk Detection (Prediction)} are the projects that predict the risk
of recidivism, and risk of corruption in the public procurement process (e.g., 
\href{https://github.com/alan-turing-institute/DSSG19-DNCP-PUBLIC}{alan-turing-institute/DSSG19-DNCP-PUBLIC}).
\end{inparaenum}

We also see projects that serve applications in various fields ranging from healthcare to the legal domain. Here we discuss some such application fields with respective repository examples:
\begin{itemize}
    \item Business: These use-cases contain projects that provide support to business and financial needs, such as deciding about credit approval, predicting financial solvency, and developing advertisement strategies. For example, the project \href{https://github.com/kratikanaskulwar/Mitacs-Verafin-Internship}{kratikanaskulwar/Mitacs-Verafin-Internship} audits bias in fraud detection machine learning model's prediction. 
    \item Legal: These use-cases contain projects that are related to legal rules, regulations, and law enforcement. Sometimes legal matters can be discriminatory towards certain groups of people. For example, the project \href{https://github.com/Caramel96/AIF360_Bias_Mitigation-police-stop-and-search-project}{Caramel96/AIF360\_Bias\_Mitigation-police-stop-and-search-project} explores discrimination in police stop and search strategy based on the race of an involved person.
    \item Health: These use-cases contain projects that are related to both physical and mental health. For example, the project  \href{https://github.com/drskaf/EHR-Prediction-Model-with-Tensorflow}{drskaf/EHR-Prediction-Model-with-Tensorflow} explores demography-based bias in calculating expected hospitalization time.
    \item Public service: These use cases contain projects that provide services to the public, including policy violation, and traffic handling. The project \href{https://github.com/AnaTorresR/DPA-food_inspections}{AnaTorresR/DPA-food\_inspections} explores fairness within the process of restaurant inspection service of a city.
\end{itemize}

\begin{figure}
\subfloat [By generic vs non-generic activity types]
      {
      \centering
      \includegraphics[scale=.55]{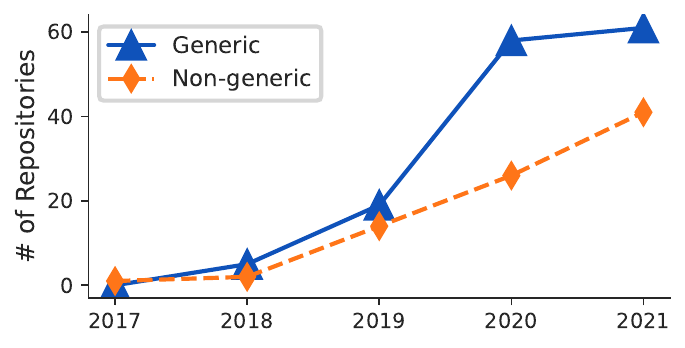}
      \label{fig:usecase-domain-trends}
      }  
\subfloat [By activity types under the non-generic use cases]
      {
      \centering
      \includegraphics[scale=.55]{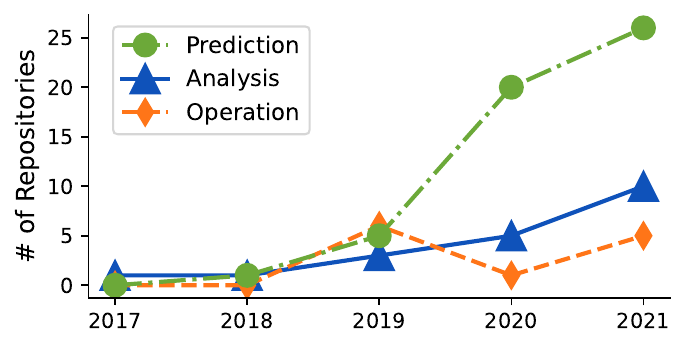}
      \label{fig:usecase-activity-trends-nongeneric}
      }  \\      
      \caption{The evolution of the identified use cases based on the creation of the corresponding GitHub repositories.}
\vspace{-3mm}
\label{fig:usecase-trends}    
\end{figure}

\fig\ref{fig:usecase-trends} provides insights into how these fairness-related repositories were growing over time. 
\fig\ref{fig:usecase-domain-trends} shows that both generic and non-generic repositories are increasing, and we see a more consistent growth trend for the non-generic repositories. Moreover, \fig\ref{fig:usecase-activity-trends-nongeneric} suggests that for all non-generic activities, the number of repositories is increasing significantly and the growth trend in the prediction-related ML use-cases is more rapid. 
This observation may indicate a growing demand for bias detection and mitigation in machine learning applications.

However, we note that these trends are presented without a baseline relative to the overall growth of machine learning repositories, which has also expanded substantially in recent years (e.g., the rapid adoption of PyTorch\footnote{\href{https://www.star-history.com/\#pytorch/pytorch\&Date}{star-history.com/\#pytorch/pytorch\&Date}}). Therefore, the proportional popularity of fairness-related repositories within the broader ML landscape may be stable rather than increasing.
Notably, generic learning and exploration use cases show the highest counts over time, which suggests an ongoing interest in educational and exploratory materials on fairness. While this could indicate that such materials are already abundant, continued refinement and dissemination of high-quality resources may still benefit practitioners and researchers seeking to integrate fairness into machine learning workflows.


\begin{rqbox}{RQ1 Summary} 
The use-cases of bias detection and mitigation APIs are diverse. In our small,
but carefully curated, dataset of 204 projects, we found 17 unique use-cases
that have been developed both for learning purposes and solving real-world issues, where use cases solving real-world issues utilize three different activities. Also, among the observed use cases, some are 
related to various types of services (e.g., health, legal, and business), and others are related to the model-related operations from the fairness APIs (e.g., ML model deployment).
Our results show developers/users devote significant effort to learning the usage of fairness APIs (observed via the over 70\% generic use cases) 
At the same time, we observe that the usage of fairness APIs for real-world non-generic use cases is steadily growing.

\end{rqbox}

\section{Bias Detection and Mitigation Approaches Reused from the Fairness APIs (RQ2, RQ3)}\label{sec:rq-detection-mitigation}

While analyzing the fairness use cases in RQ1 (\sec\ref{sec:lib_usecases}), we 
observed that fairness APIs contain various types of metrics
to detect bias and techniques/strategies to mitigate those biases. 
In this section, we discuss how such bias
detection and mitigation approaches are utilized in the studied ML software systems. We answer two research questions:
\begin{enumerate}[label=\textbf{RQ\arabic{*}}., start=2]
    \item How are the biases detected in the use-cases?
    \item How are the biases mitigated in these use-cases?
\end{enumerate}

\subsection{RQ2. How are the biases detected in the use-cases?}\label{sec:rq-bias-detection-strategies}

\subsubsection{Motivation} Numerous bias detection metrics have been proposed by
the research community, and many of them have already been implemented. For
example, the AI Fairness 360 library has implemented about 70 types of fairness
metrics \cite{AIFairness3602018} as Python APIs. The objective of this research
question is to explore how real-world software projects apply these
libraries for detecting bias across the use cases we identified in \sec\ref{sec:lib_usecases}.

\subsubsection{Approach}\label{sec:rq2-approach} In \sec\ref{sec:lib_usecases}, we found that around 29.4\% of the ML software 
repositories in our dataset contained use cases that are non-generic, i.e., the use cases are applied for real-world 
applications where bias and fairness concerns need to be addressed. Given the relative importance of those non-generic use cases over the 
generic use cases, in this research question, we specifically focus on the bias detection approaches used in the non-generic use cases.
As such, we analyzed 60 ML software repositories that contained non-generic use cases.
 
Each GitHub repository in our dataset for non-generic use cases has used at least one of the fairness API libraries
listed in Table~\ref{tab:libraries}. Each of these API libraries has
multiple methods. Each method can use one or more fairness metrics to detect bias. 
It is possible that a project in our final set did not
use all the available methods from a library. To understand how an API is used to detect bias in a given use case scenario, 
we first listed all the
available methods within our selected API libraries. For most API libraries, it was done by
studying their well-written documentation that listed all the available
detection and mitigation methods. For libraries without sufficient documentation,
we studied their publication sources (i.e., published research papers), GitHub
README files, and inspected the project's source code when necessary.

Before we were able to analyze how the available APIs are being used across the selected set of repositories, we first needed to collect all the API calls that had been done utilizing the library imports within our selected repositories. For this, we used the AST analysis approach utilizing the AST module from Python standard library. During this AST analysis approach, we considered various possible ways of importing API libraries and performing the API/function calls. For example, we considered both globally scoped module imports as well as locally scoped imports, calling APIs using techniques other than direct API calls (e.g., assigning the API into an attribute and later using that attribute for API calls, etc.). Considering such cases helped us to lessen the number of missed/undetected valid API calls. We then used the \textit{substring} matching approach to match the extracted method names from these selected software projects' source code against our prepared list of available APIs (here, detection methods).

Once the methods are identified, we manually determine the fairness metrics that are used from the methods 
in a GitHub repo. We then find the usage of these metrics within our selected GitHub repositories.
Finally, we classify the bias detection metrics to understand their distribution across 
these repositories and the identified use cases. This classification primarily
focuses on the cardinality of the involved parties---i.e., whether we are
measuring fairness among population groups or individuals. 
The classification uses the five types of bias detection metrics we discussed in \sec\ref{sec:background-bias-detection}.

To note, it is possible that one repository used several types of bias detection or mitigation methods. For such cases, when we counted the number of repositories for a given detection or mitigation method, the same repository was counted separately for each of the methods. 


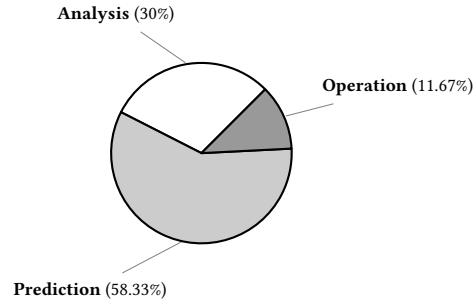
\begin{figure}[t]
      \centering\begin{tikzpicture}[scale=0.40]-
      \tikzstyle{every node}=[font=\footnotesize]
    \pie[
        /tikz/every pin/.style={align=center},
        text=pin, number in legend,
        explode=0.0, rotate=45,
         color={black!0, black!20, black!40, black!60},
        ]
        {
            30.0/\bf{Analysis} (30\%),
            58.3/\bf{Prediction} (58.33\%),
            11.7/\bf{Operation} (11.67\%)
        }
    \end{tikzpicture}
      \caption{ Distribution of the use-cases of the repositories that used fairness metrics}
\vspace{-3mm}
\label{fig:detection_dist}    
\end{figure}

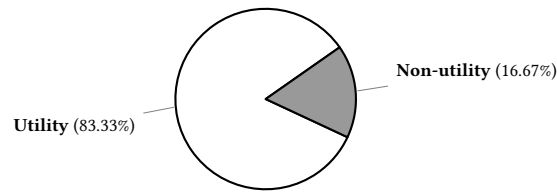
\begin{figure*}[t]
      \centering\begin{tikzpicture}[scale=0.40]-
      \tikzstyle{every node}=[font=\footnotesize]
    \pie[
        /tikz/every pin/.style={align=center},
        text=pin, number in legend,
        explode=0.0, rotate=35,
         color={black!0, black!40},
        ]{
            83.33/\bf{Utility} (83.33\%),
            16.67/\bf{Non-utility} (16.67\%)
        }
    \end{tikzpicture}
      \caption{Distribution of the fairness metrics across all the 60 repositories that contained non-generic use cases.}
\label{fig:detection_metric_types}
\end{figure*}


\subsubsection{Results}\label{sec:rq-bias-detection-results} 
\fig\ref{fig:detection_dist} shows the distribution of the activity types that used bias detection APIs. 
Prediction (58.33\%) is the activity type that contains most of the used bias detection metrics, followed by Analysis, and finally by Operation. 



Most of the projects (83.33\% ) in our dataset detected bias with the adoption and
modification of the utility metrics. Group fairness metrics were directly
applied by 16.67\% of the projects. Other categories, such as bias detection for
subgroups and individuals, are non-existent.

Table~\ref{tab:detection_usecases} shows how these bias detection APIs have been
used across the non-generic applications we found in \sec\ref{sec:lib_usecases}. For each metric type, we mention how many of the non-generic repositories contain usage of this metric and report the activity distribution of these repositories. Here we see that utility metrics are very prevalent in prediction-related use cases. However, the group metrics have been used evenly across all different activity types of non-generic applications.  


\begin{table}[t]
    \centering

    \caption{ Distribution of the bias detection metrics across the different use-cases in our studied GitHub repositories. Here, \#R = number of GitHub repositories.}  
    
    \label{tab:detection_usecases}
    \begin{tabular}{lll}
    \toprule
    \textbf{Metrics} 	&  \textbf{Non-generic Use-case}  &			\textbf{\#R} \\
    \midrule
\textbf{Utility} & Prediction & 29\\
Non-generic: 50 
& Analysis & 16  \\
 & Operation & 5\\
\cmidrule {1-3}
\textbf{Group} &  &     \\
\cmidrule {2-3}
\textit{Separation}  & Analysis & 4 \\
Non-generic: 5 
& Prediction & 1 \\
\cmidrule {2-3}
\textit{Independence}  & Operation & 7 \\
Non-generic: 4 
& Analysis & 3 \\
 & Operation & 1 \\
\cmidrule {2-3}
\bottomrule
\end{tabular}%
\end{table}

In order to have more insights into how developers implement bias detection libraries, we provide two concrete examples of bias detection below.

\textbf{\textit{Metropolitan police}}: This use-case is implemented in a GitHub repo
\href{https://github.com/danladishitu/CE888-7-SP}{danladishitu/CE888-7-SP}, 
where the developer wanted to find out if there is any bias based on
the ethnicity of a person when metropolitan police had to use force in certain
situations. The implementation involves the following steps:

\begin{inparaenum}[(1)]
\item Importing the relevant bias-related methods from the \texttt{aif360} API
(Listing \ref{lst:usecase1_import}).
\item Preparing the dataset for applying bias detection metrics or mitigation
algorithms (Listing \ref{lst:usecase1_dataset}). Here the developer used a
dataset with binary label and ethnicity as the protected attribute.
\item Detecting bias on the classifier's prediction (Listing
\ref{lst:usecase1_detection}). Here, all three kinds of group bias
detection metrics such as sufficiency (True positive rate, True negative rate),
separation (Average odd difference), and independence (Disparate impact) were
used to calculate group fairness in the classifier's output. In \sec\ref{sec:background-bias-detection}, we provided more details about these fairness metrics.
\end{inparaenum}

\begin{lstlisting}[caption={Metropolitan police use-case: API imports},label={lst:usecase1_import},basicstyle=\footnotesize]
from aif360.datasets import BinaryLabelDataset 
from aif360.metrics import ClassificationMetric
from aif360.explainers import MetricTextExplainer  
from aif360.algorithms.preprocessing import Reweighing
\end{lstlisting}

\begin{lstlisting}[caption={Metropolitan police use-case: Preparing dataset for applying the API},label={lst:usecase1_dataset},basicstyle=\footnotesize]
bld = <@\textcolor{black}{\colorbox{yellow}{BinaryLabelDataset}}@>(df=pd.concat((x, y_true),
            axis=1),
            label_names=['Impact Factor: Possesion of a weapon'],
            protected_attribute_names=['SubjectEthnicity'],
            favorable_label=1,
            unfavorable_label=0)
privileged_groups = [{'SubjectEthnicity': 1}]
<@\textcolor{black}{\colorbox{yellow}{unprivileged\_groups}}@> = [{'SubjectEthnicity': 0}]
\end{lstlisting}

\begin{lstlisting}[caption={Metropolitan police use-case: Detecting bias using various classification metrics},label={lst:usecase1_detection},basicstyle=\footnotesize]
valid_metric =  <@\textcolor{black}{\colorbox{yellow}{ClassificationMetric}}@>(bld, bld_preds, 
            unprivileged_groups=unprivileged_groups,
            privileged_groups=privileged_groups)

balanced_accuracy.append(0.5 * (valid_metric. <@\textcolor{black}{\colorbox{yellow}{true\_positive\_rate}}@>()
            + valid_metric. <@\textcolor{black}{\colorbox{yellow}{true\_negative\_rate}}@>()))
avg_odd_diff.append(valid_metric. <@\textcolor{black}{\colorbox{yellow}{average\_odds\_difference}}@>())
disp_impact.append(np.abs(valid_metric. <@\textcolor{black}{\colorbox{yellow}{disparate\_impact}}@>() - 0.5))
\end{lstlisting}

\textbf{\textit{Hospitalization time}}: We can find another example of usage of bias detection approaches within the GitHub repo 
\href{https://github.com/drskaf/EHR-Prediction-Model-with-Tensorflow}{drskaf/EHR-Prediction-Model-with-Tensorflow}. The primary aim of this project is to estimate the hospitalization time of the patients. In this prediction of hospitalization time, along with other relevant attributes, the developers also used demographic information (such as race and gender) as training attributes. So, in order to verify whether the prediction of their trained model is biased towards any specific demography group, the developers used the group bias detection API from the Aequitas library for performing this inspection. 
We can divide their implementation into three sequential segments: 
\begin{inparaenum}[(a)]
\item Importing the appropriate APIs (Listing \ref{lst:usecase2_import}).
\item Calculating group fairness considering race and gender as sensitive attributes (Listing \ref{lst:usecase2_detail}).
\item Plotting the output of the fairness metrics for better understanding (Listing \ref{lst:usecase2_plotting}). 
\end{inparaenum}
\begin{lstlisting}[caption={Hospitalization time use-case: API imports},label={lst:usecase2_import},basicstyle=\footnotesize]
from aequitas.preprocessing import preprocess_input_df
from aequitas.group import Group
from aequitas.plotting import Plot
from aequitas.bias import Bias
from aequitas.fairness import Fairness
\end{lstlisting}

\begin{lstlisting}[caption={Hospitalization time use-case: Calculating and visualizing group fairness},label={lst:usecase2_detail},basicstyle=\footnotesize]
ae_subset_df = pred_test_df[['race', 'gender', 'score', 'label_value']]
ae_df, _ = <@\textcolor{black}{\colorbox{yellow}{preprocess\_input\_df}}@>(ae_subset_df)
g = Group()
xtab, _ = g.get_crosstabs(ae_df)
absolute_metrics = g.<@\textcolor{black}{\colorbox{yellow}{list\_absolute\_metrics}}@>(xtab)
clean_xtab = xtab.fillna(-1)
aqp = Plot()
b = Bias()

# Reference Group Selection
# test reference group with Caucasian Male
bdf = b.<@\textcolor{black}{\colorbox{yellow}{get\_disparity\_predefined\_groups}}@>(clean_xtab, 
                    original_df=ae_df, ref_groups_dict={'race':'Caucasian', 'gender':'Male'}, 
                    alpha=0.05, check_significance=False)

f = Fairness()
fdf = f.<@\textcolor{black}{\colorbox{yellow}{get\_group\_value\_fairness}}@>(bdf)
\end{lstlisting}

\begin{lstlisting}[caption={Hospitalization time use-case: Plotting fairness metrics},label={lst:usecase2_plotting},basicstyle=\footnotesize]
# Race and Gender Bias Analysis for Patient Selection
# Plot two metrics

# Is there significant bias in your model for either race or gender?
p = aqp.<@\textcolor{black}{\colorbox{yellow}{plot\_group\_metric\_all}}@>(xtab, metrics=['tpr', 'ppr'], ncols=2)

# Fairness Analysis Example - Relative to a Reference Group
# Reference group fairness plot
fpr_disparity = aqp.<@\textcolor{black}{\colorbox{yellow}{plot\_disparity}}@>(bdf, group_metric='fpr_disparity', 
                                        attribute_name='race')
\end{lstlisting}

\begin{figure}[h]
\centering
\includegraphics[width=0.65\columnwidth,keepaspectratio]{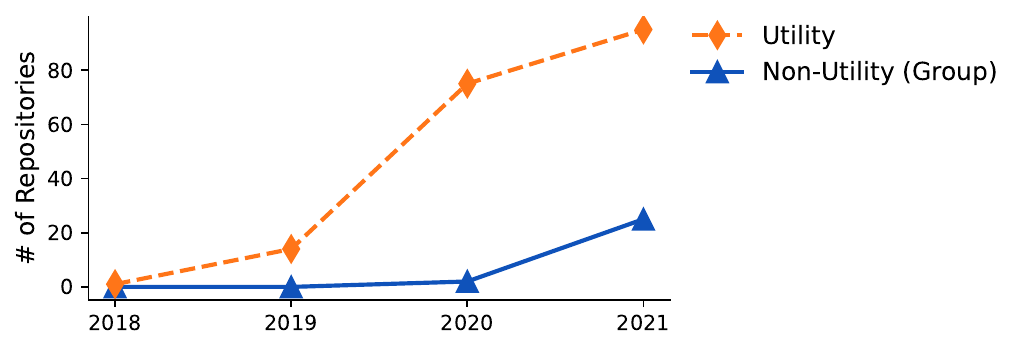}
\caption{Evolution of the bias detection metric categories based on the integration of the APIs on corresponding GitHub repositories.}
\vspace{-3mm}
\label{fig:detection-trends}
\end{figure}

We are also interested to learn how the fairness metrics used within our selected repository
set evolved over time. To perform such trend analysis, we first collected the source code of all selected repositories along with the full git source tree of these repositories. We need the full git source tree since we want to see which APIs from the available APIs (from our selected API libraries) had been used within a particular year. So, for each repository, we considered one commit per year starting from the repository creation date. To be more specific, we considered the last commit of each year as the final state of that repository for that year. Next, we performed the API detection steps (described in \sec\ref{sec:rq2-approach}) on those selected sets of commits by individually checking out the source code of the repository at those selected sets of commits. Finally, after performing these detection steps, for each repository, we had a list of used detection APIs for each of the years of the repository's development. We then used this collected information to perform the trend analysis. Now we describe our findings below.

\fig\ref{fig:detection-trends} shows that the application
of the utility and group fairness metrics has increased over time. The growth in the use of utility metrics slowed a bit in 2021 compared to in 2020. In contrast, the group fairness metrics
have been used monotonously since 2018, probably because the API users are
becoming more aware of the available fairness metrics and have started to use metrics
that are specific to their needs. As we didn't find any use of the other three fairness metrics (i.e., Subgroup, Individual, Group-Individual) within the filtered set of non-generic repositories, there were no trend plots for these metric types.

\begin{rqbox}{RQ2 Summary} 
Although there are five categories of fairness metrics, all of the non-generic software repositories use only the utility or the group fairness metrics for bias detection. Fairness for subgroups and individuals has been largely ignored, which could be devastating in systems where bias may incur at finer levels. Future studies should explore the challenges of developing and adopting these finer-level bias detection metrics.
\end{rqbox}


\subsection{RQ3. How are the biases mitigated in these use-cases?}\label{sec:rq-bias-mitigation-strategies}

\subsubsection{Motivation} Similar to the bias detection approaches, numerous bias mitigation approaches have been proposed and developed. 
The objective of this research question is to explore how real-world software projects apply these libraries for bias mitigation. 

\subsubsection{Approach}
Similar to RQ2, we focus on the 60 GitHub repositories that utilized the fairness APIs for 
non-generic use cases. Each GitHub repository in the final set has used at least one of the fairness API libraries
listed in Table~\ref{tab:libraries}. Seven out of the 13 API libraries offer methods to mitigate bias. Each of these API libraries has
multiple methods to mitigate the biases. It is possible that a project in our final set did not
use all the available methods from an API. We, therefore, first identify 
which mitigation methods of an API are used in each of our studied GitHub repos. 

Similar to our analysis of bias detection methods (discussed in \sec\ref{sec:rq2-approach}), 
we consulted the API documentation, GitHub README files, and corresponding publication sources 
to collect the names of mitigation methods in each API. We then used
the \textit{substring} approach to match the collected method names against the method names imported from the API within the source of each GitHub repo. 

\tbl\ref{tab:mitigation_approach} shows the bias mitigation approaches that have
been implemented as APIs in our selected libraries. Most of these APIs (written
in \textit{italic}) have been published as research articles. A few of the
mitigation approaches (e.g., Correlation remover,  Upsampling) are, however,
provided out of the box by the libraries to mitigate bias. Evidently, AI
Fairness 360, and EthicML are the most prevalent; they have most of the
mitigation approaches implemented. On the contrary, other libraries have
implementation only for a few selected mitigation approaches.
For example, BlackBoxAuditing has implementation only for the disparate impact
remover approach. Interestingly, Six of the 13 fairness libraries (i.e.,
Aequitas, Fairness Indicators, FairML, Transparent AI, Failens, and Fair-forensics) do not have any mitigation approach implemented; they implemented only bias detection
approaches, or only contain biased datasets for experimentation.
\begin{table*}[t]
    \centering
    \caption{Bias mitigation approaches that are implemented as APIs in our selected API libraries}
    \label{tab:mitigation_approach}
    \scalebox{0.8}{
    \begin{tabular}{llr}
    \toprule
    \textbf{Mitigation Approaches}  										&  \textbf{Introduced By}  &					 \textbf{Implemented in Fairness Libraries}  \\ \midrule
	\textit{Disparate impact remover}  & Feldman et al. (2015)  \cite{feldmanCertifyingRemovingDisparate2015}  & AI Fairness 360, BlackBoxAuditing, Fairness Comparison \\
    \textit{Learning fair representations} & Zemel et al. (2013)  \cite{zemelLearningFairRepresentations2013}	&	AI Fairness 360, EthicML \\
    \textit{Optimized preprocessing} & Calmon et al. (2017) \cite{calmonOptimizedPreProcessingDiscrimination2017}	& AI Fairness 360 \\
    \textit{Reweighing} & Kamiran et al. (2012) \cite{kamiranDataPreprocessingTechniques2012a} & AI Fairness 360, EthicML \\
    \textit{Massaging} & Kamiran et al. (2012) \cite{kamiranDataPreprocessingTechniques2012a} & EthicML \\
    Correlation remover	&	Fairlearn (Library) \cite{FairlearnDocumentationFairlearn} & Fairlearn \\
    \textit{Adversarially learned fair representations}	&	Beutel (2017) \cite{beutelDataDecisionsTheoretical2017}	&	EthicML	\\
    Upsampling	&	EthicML (Library) \cite{EthicMLDocumentationMitigation}	&	EthicML	\\
    \textit{Variational Fair Auto-Encoder}	&	Louizos et al. (2017) \cite{louizosVariationalFairAutoencoder2017}	&	EthicML	\\
    Relabelling	&	ThemisML (Library) \cite{bantilanThemisMLSource2017}	& ThemisML	\\
    \textit{Discrimination-Free Classification}	&	Calders et al. (2010) \cite{caldersThreeNaiveBayes2010}	&	Fairness Comparison \\
    \textit{Adversarial debiasing} & Zhang et al. (2018) \cite{zhangMitigatingUnwantedBiases2018}	&	AI Fairness 360	\\
    \textit{Gerry Fair classifier}	&	Kearns et al. (2018) \cite{kearnsPreventingFairnessGerrymandering2018}	&	AI Fairness 360, Gerryfair	\\
    \textit{Meta Fair classifier}	&	Celis et al. (2020) \cite{celisClassificationFairnessConstraints2020}	&	AI Fairness 360	\\
    \textit{Prejudice remover}	&	Kamishima et al. (2012)	\cite{kamishimaFairnessawareClassifierPrejudice2012}	&	AI Fairness 360, EthicML \\
    \textit{Exponentiated gradient reduction}	&	Agarwal et al. (2018) \cite{agarwalReductionsApproachFair2018}	&	AI Fairness 360, Fairlearn, EthicML \\
    \textit{Grid search reduction}	&	Agarwal et al. (2018) \cite{agarwalReductionsApproachFair2018}	&	AI Fairness 360, Fairlearn \\
	\textit{Distributionally-robust optimization}	&	Hashimoto et al. (2018) \cite{hashimotoFairnessDemographicsRepeated2018}	&	EthicML \\
    \textit{CORELS (Certifiably Optimal RulE ListS)}	&	Angelino et al. (2018) \cite{angelinoLearningCertifiablyOptimal2018}	&	EthicML \\
    \textit{Learning Classification without Disparate Mistreatment}	&	Zafar et al. (2017) \cite{zafarFairnessDisparateTreatment2017}	&	EthicML, Fairness Comparison \\
    \textit{Calibrated equalized odds}	&	Pleiss et al. (2017) \cite{pleissFairnessCalibration2017}	&	AI Fairness 360	\\
    \textit{Equalized odds} & Hardt et al. (2016) \cite{hardtEqualityOpportunitySupervised2016a} &	AI Fairness 360, Fairlearn, EthicML \\
    \textit{Reject option classification}	&	Kamiran et al. (2012) \cite{kamiranDecisionTheoryDiscriminationAware2012}	&	AI Fairness 360, ThemisML \\
    DP flip	&	EthicML (Library) \cite{FairlearnDocumentationFairlearn}	&	EthicML \\
    
    \bottomrule
\end{tabular}
}
\end{table*}

In order to have insights into how these mitigation approaches work, and how
they are used in the wild, we first merge these mitigation approaches into
higher-level more generic approaches. For example, we merge the
\textit{Correlation remover} and the \textit{Desperate impact remover}
mitigation APIs to higher-level \textit{data transformation}, because they
internally use different data transformation techniques to mitigate bias within
a dataset. These generic approaches were then merged into the three highest-level approaches, in accordance with previous
studies~\cite{mehrabiSurveyBiasFairness2021,bellamyAIFairness3602018}, which
are known as pre-processing, in-processing, and post-processing.  Two
authors went through the source materials of these mitigation approaches (e.g., original publications, tutorials, sample implementations) to
prepare this higher-level categorization. Two authors first independently went
through these source materials and then worked together until they reach an
agreement about the categorization of these approaches. This process took
$\sim$20 hours of online discussion.

\subsubsection{Results}\label{sec:rq-bias-mitigation-results}

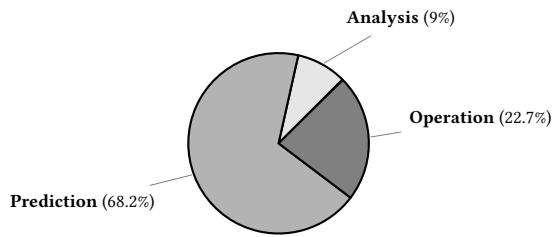
\begin{figure}[t]
      \centering\begin{tikzpicture}[scale=0.40]-
      \tikzstyle{every node}=[font=\footnotesize]
    \pie[
        /tikz/every pin/.style={align=center},
        text=pin, number in legend,
        explode=0.0, rotate=45,
         color={black!10, black!30, black!50, black!70},
        ]
        {
            9/\bf{Analysis} (9\%),
            68.2/\bf{Prediction} (68.2\%),
            22.7/\bf{Operation} (22.7\%)
        }
    \end{tikzpicture}
      \caption{Distribution of use-cases of the repositories that used the bias mitigation approaches}
\vspace{-3mm}
\label{fig:mitigation_dist}    
\end{figure}

\fig\ref{fig:mitigation_dist} shows that 68.2\% of the repositories that applied biased mitigation are within the Prediction activity type, followed by Operation (22.7\%) and Analysis (9\%). If we compare with \fig\ref{fig:detection_dist} where we discussed a similar type of distribution for the detection metrics, we see that Prediction activity tops in applying both detection metrics and mitigation algorithms. However, we also see that repositories with Analysis activity focused more on detection than on applying mitigation approaches. 


\begin{figure}[t]
\centering
\includegraphics[scale=0.6]{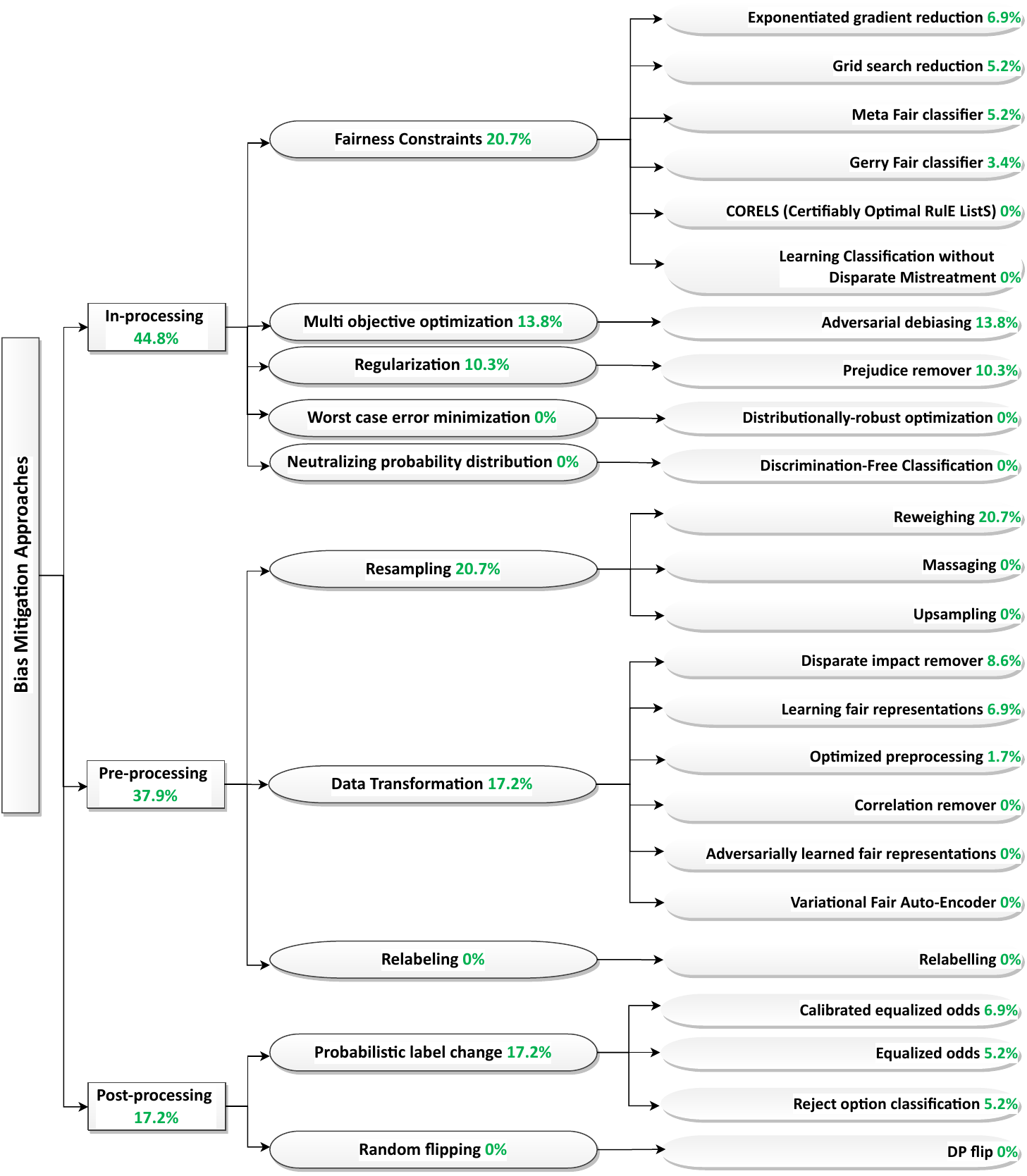}
\caption{Classification of bias mitigation approaches.}
\label{fig:mitigation_approach}
\end{figure}

\fig\ref{fig:mitigation_approach} presents the distribution of overall used
mitigation approaches.
We see that In-processing is the most common approach to bias mitigation in our GitHub repositories (44.8\%), where the Fairness constraints (20.7\%) is the most dominant in-processing approach. Also, only 13.8\% of the projects followed the multi-objective optimization approach.

Along with that, we see the worst-case error minimization approach along with Neutralizing probability distribution were not adopted by any of our repositories. 

Next, we see 37.9\% of our GitHub projects adopted different pre-processing approaches for bias mitigation. Resampling is the most common pre-processing approach (20.7\% GitHub repositories), followed by Data transformation, the second most prevalent pre-processing approach (17.2\%). On the other hand, the learning fair representations approach~\cite{zemelLearningFairRepresentations2013}) has been applied by 6.9\% of the repositories. 

Finally, we find that 17.2\% of our subject repositories used post-processing
mitigation all of which came from the usage of the Probabilistic label change
approach, as the other post-processing approach, Random flipping, was not used by
any of the studied repositories.

Table~\ref{tab:mitigation_usecases} shows how different mitigation approaches
were used across different use-cases found in RQ1. Mitigation approaches are
indeed used across very different use-cases (i.e., different activities and
different domains) to reduce bias and improve the overall fairness of the
systems. We find that different repositories with the same
use case can apply different bias mitigation approaches. For example, the
repositories for credit approval (prediction activity in
Table~\ref{tab:mitigation_usecases}) have applied all three highest-level
mitigation approaches, and with different algorithms: pre-processing (with
resampling, and data transformation),  in-processing (with fairness constraints,
and regularization), and post-processing (with probabilistic label change). This
observation is also true for disease prediction use case as well.

\begin{table}[htb]
    \centering
    \caption{Usage of mitigation approaches in different use-cases. Here \#R = Number of GitHub repositories}
    \label{tab:mitigation_usecases}
    \begin{tabular}{lll}
    \toprule
    \textbf{Mitigation}  &	\textbf{Non-generic Use-case}  &  \textbf{\#R} \\ \toprule
\bf{Pre-processing }  &  &   \\
\textit{Resampling}  & Prediction & 10 \\
Non-generic: 11
& Operation & 1 \\
\cmidrule {2-3}
\textit{Data transformation} & Prediction & 3 \\
Non-generic: 5 
& Analysis & 1 \\
  & Operation & 1 \\
\midrule
\bf{In-processing}  &  &   \\
\textit{Fairness constraints}  & Prediction & 6 \\
Non-generic: 11 
& Operation & 4 \\
 & Analysis & 1 \\
\cmidrule {2-3}
\textit{Multi-objective optimization} & Prediction & 5 \\
Non-generic: 8
& Operation & 2 \\
  & Analysis & 1 \\
\cmidrule {2-3}
\textit{Regularization} & Prediction & 5 \\
Non-generic: 6  
& Operation & 1 \\
\midrule
\bf{Post-processing}  &  &  \\
\textit{Probabilistic label change} &  Prediction & 6 \\
Non-generic: 8 
& Analysis & 1 \\
 & Operation & 1 \\
\bottomrule
\end{tabular}%
\end{table}


In Listing \ref{lst:usecase1_mitigation}, we show how bias was mitigated, using
the \textit{Reweighing} pre-processing algorithm
\cite{kamiranDataPreprocessingTechniques2012a}, in the metropolitan police use
case described in \sec\ref{sec:rq-bias-detection-strategies}.
After applying the mitigation algorithm the repository developers generally
verify whether the situation improved or not by re-calculating bias detection
metrics. For example, here \textit{BinaryLabelDatasetMetric} has been used for
recalculating the dataset metric.

\begin{lstlisting}[caption={Metropolitan police use-case: Mitigating bias and recalculating bias metrics},label={lst:usecase1_mitigation},basicstyle=\footnotesize]
rw =  <@\textcolor{black}{\colorbox{yellow}{Reweighing}}@>(unprivileged_groups=unprivileged_groups,
                privileged_groups=privileged_groups)
train_pp_bld_f = rw.fit_transform(train_pp_bld)

metric_train_bld = <@\textcolor{black}{\colorbox{yellow}{BinaryLabelDatasetMetric}}@>(train_pp_bld_f,
                  unprivileged_groups=unprivileged_groups,
                    privileged_groups=privileged_groups)
\end{lstlisting}

\begin{figure}[htb]
\centering
\subfloat[Evolution of different types of mitigation approaches at the highest level.]
      {
      \centering
      \includegraphics[scale=.55]{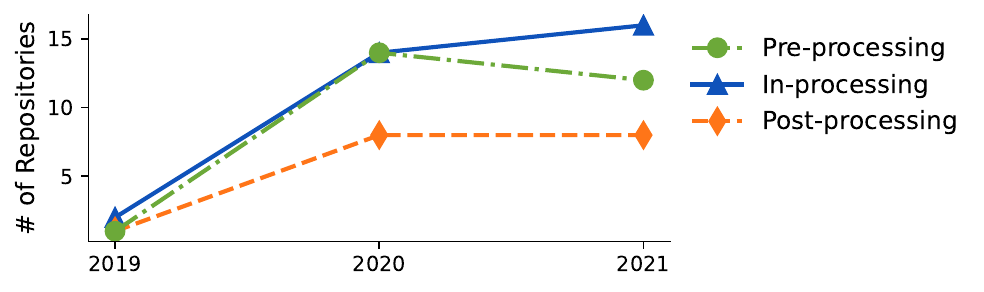}
      \label{fig:mitigation-type-trends}
      } \\
\subfloat [Evolution of different types of mitigation algorithms.]
      {
      \centering
      \includegraphics[scale=.55]{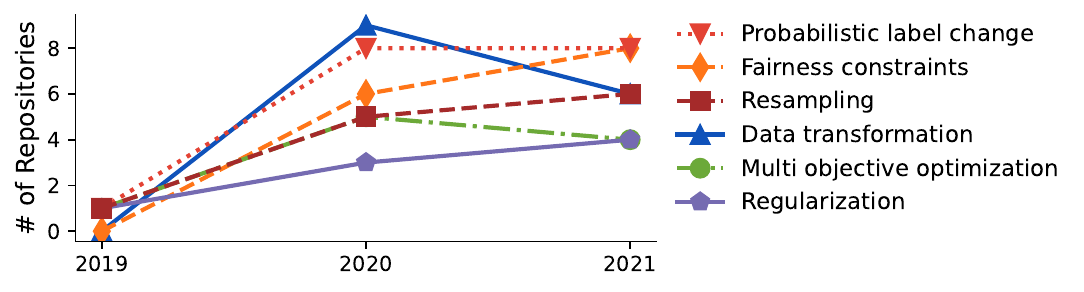}
      \label{fig:mitigation-subtype-trends}
      } 
      \caption{Evolution of bias mitigation approaches based on the integration of the APIs on corresponding GitHub repositories.}
\vspace{-3mm}
\label{fig:mitigation-trends}    
\end{figure}

\fig\ref{fig:mitigation-trends} shows how the use of different types of bias
mitigation approaches evolved over the years. 
To perform this trend analysis, we used a similar approach that we followed for analyzing the trends of usage of the detection APIs (described in \sec\ref{sec:rq-bias-detection-results}). 
\fig\ref{fig:mitigation-type-trends} shows the trend at the highest level (i.e.,
Pre-processing, In-processing, and Post-processing). Pre-processing---removing
bias from training data---has always been the most dominant mitigation approach.
In-Processing, although less popular at the beginning, has become the most
popular approach over time. On the other hand, even though Post-processing has been the
least popular of the three, one of its algorithms, probabilistic label change, has always been
the most dominant bias mitigation approach until 2021
(Figure~\ref{fig:mitigation-subtype-trends}) when steadily rising fairness constraints approach took over its place.
In general, we observe an overall increase in using bias mitigation in ML
software. This implies the growing demand for fairness improvement and bias mitigation in ML systems.

\begin{rqbox}{RQ3 Summary} 
The primary implications of our observations are two.  i) Demand for bias mitigation in machine learning software is increasing over time. This is an important observation for the research community who, therefore, should focus on building more bias mitigation approaches and should investigate more on solving the current mitigation-related problems, such as making a bias mitigation approach to be transferable~\cite{jinTransferabilityBiasMitigation2021}. ii) Although the in-processing approaches are the most common bias mitigation approach, for the same use cases bias can be mitigated by multiple approaches. As such, there is a need to study which mitigation approach is better in a given scenario, which may act as a guideline for the developers. 

\end{rqbox}

\section{Challenges of Reusing and Designing the Fairness APIs (RQ4)} \label{sec:challenges-bias-apis}


\subsection{Motivation} Our findings in RQ1 - RQ3 show that the fairness API libraries are used to support diverse use case scenarios to detect and mitigate biases. 
We also find that a considerable number of the applications focus on tutorials or generic usage, which could denote that the users of the API libraries need to spend time and effort 
learning how to use the API libraries before they apply those to real-world use cases (e.g., healthcare). As such, it is important to understand whether the API libraries can be improved and/or 
adapted to support and increase their usage. To understand how the API libraries can be improved, we first need to understand the types of challenges the developers and users of the API libraries face while using and developing the API libraries. This insight can help the developers of the API libraries to better design their API libraries along with the key topic areas of challenges.
In summary, we explore the following research question:

\nd\bf{RQ4. What are the topics found in the issues that developers reported while using the API libraries?}

\subsection{Approach} To understand the challenges the API developers of fairness libraries
face, we applied topic modeling on the 4212 GitHub issue discussions (989 issue reports and 3223 associated
comments) of these libraries that we collected in \sec\ref{sec:collect-issue-comments}. 


To identify the topics in the issues reported by developers, we analyzed both the issue descriptions and the associated comments. Since discussions within a single issue often diverge into unrelated subjects, we treated issue descriptions and comments as separate units. Initially, we merged each issue's description and all of its comments into a single document sample. However, this approach led to inconsistent topic modeling results, where irrelevant or unrelated topics were extracted due to the mixed nature of the content. This is because a single issue often involves discussions on multiple distinct topics. For example, the issue \href{https://github.com/fairlearn/fairlearn/issues/265}{fairlearn/issues/265} includes discussions ranging from bias in APIs to troubleshooting, installation problems, library usage, and version compatibility. Aggregating all these discussions into one document resulted in overlapping and incoherent topic assignments. Conversely, treating each comment as a separate document does not necessarily produce entirely distinct topics; comments that focus on similar themes naturally cluster under the same topic.


Our approach to topic modeling is similar to~\cite{scocciaChallengesDevelopingDesktop2021}, where
Scoccia \textit{et al.} analyzed the challenges in developing desktop web apps
using the issue reports from the apps' repositories that are hosted on GitHub. We apply topic modeling and analyze the output of topic modeling as follows.

First, we preprocess the issue discussions. We consider the issue titles,
the issue bodies, and the comment bodies as text fields. We then applied 
noise-reducing steps following previous
studies~\cite{uddinEmpiricalStudyIoT2021,alaminEmpiricalStudyDeveloper2021}. We
first removed all the code segments and HTML tags used to mark non-text blocks.
Then we removed stop words (i.e., ``a'', ``the'', ``this'', etc.), numbers,
non-alphabetical characters, and punctuation marks. We then applied Porter
stemming to get the roots of the words. After this step, we were left with  issue discussions due to the removal of noisy and empty discussions.

Second, we apply topic modeling to the preprocessed issue contents. Similar to
The existing studies
(e.g.,~\cite{uddinEmpiricalStudyIoT2021,alaminEmpiricalStudyDeveloper2021}), we
used the \emph{Latent Dirichlet Allocation} (LDA) algorithm
\cite{bleiLatentDirichletAllocation} provided by the MALLET library
\cite{mccallumMALLETMachineLearning}. The algorithm provides a list of topics to
group the issue discussion texts into \textit{T} topics, taking the
pre-processed texts as input. To find the optimal number of topics, we followed
the standard practice proposed by Arun \textit{et
al.}~\cite{arunFindingNaturalNumber2010}.  We have utilized the c.v. metric from
the \texttt{GenSim} package~\cite{rehurekSoftwareFrameworkTopic2010} for
calculating the coherency scores for each of these topics, and used this score
to get the optimal number of topics, as done by previous studies~
\cite{uddinAutomaticSummarizationAPI2017,uddinEmpiricalStudyIoT2021,alaminEmpiricalStudyDeveloper2021}.
As the two hyper-parameters of LDA, we used $\alpha = 50/T$ and $\beta = 0.01$
and applied MALLET LDA for $T = {5~to~50}$, incremented by 1. The highest
coherency score generated by LDA was 0.54 for $T=10$ and for $T = 20$ topics.
However, after the first two authors went through several $T < 30$ topics (e.g.,
10 and 20 topics), they found that the generated topics were quite noisy and
contained multiple types of discussions in one single topic. So the authors took
the highest coherency score where $T > 30$, and found that $T=33$ topics---with
a coherence score of 0.53, pretty close to the highest score of 0.54---leads
less noisy topics. Subsequently, 33 was selected as the optimal
number of topics. 

Third, we manually label each topic. The label of a topic summarizes the
underlying concepts of that topic. For labeling the topics, we applied an open
card sorting approach \cite{williamCardSorting2013}, which is popular in the
community for labeling LDA-generated
topics~\cite{bagherzadehGoingBigLargescale2019,uddinEmpiricalStudyIoT2021}. In
this approach, a list of top words (10 words in our case) within a topic is used
along with a list of randomly selected text samples (10 samples in our case) to
find the label of a topic. Two of the authors labeled these topics by
synchronously communicating together using Skype and Zoom meetings. They only
labeled the topics after reaching a complete agreement over the labels; in case
of a few disagreements, the third author was involved. After the first round of
labeling, we merged a number of similar topics together as they indicate
identical issues. For example, topics 21 (installation) and 4 (commands) were merged into the same topic,
\emph{Installation and shell commands}, as both of these topics discussed different
installation-related challenges. At the end of the discussion and merging
similar topics, we obtained 10 unique topics, which shows the 10 unique
challenges the developers face while developing or deploying fairness libraries.

Fourth, we build a hierarchy of the found topics. Given the fairness API libraries are used to improve the detection and mitigation of bias in 
ML software systems during the design and development of such software systems, we followed Alamin and Uddin~\cite{alalaminQualityAssuranceChallenges2021} and used the stages
of SDLC (Software Development Life Cycle) to build our higher-level topic category. For example,
\emph{Troubleshooting}, and \textit{Fixing Warning} all are
related to maintaining the implementation of the API library, so we grouped
these together under the ``Maintenance'' category. This whole process took $\sim$25 hours of discussion.

\subsection{Results}

\begin{figure}[t]
\centering
\includegraphics[scale=.55]{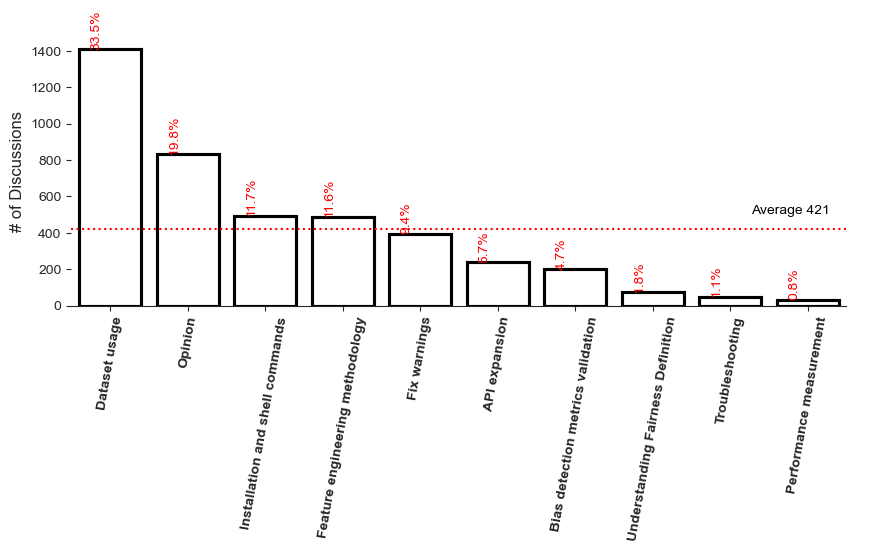}
\caption{Distribution of topics by the total number of issue discussions.}
\label{fig:topic_percentage}
\end{figure}

\fig\ref{fig:topic_percentage} shows the distribution of the 10 unique topics in our dataset of  issue discussions. 
The topics are ordered based on the total number of issue discussions assigned to each topic. 
For example, the topic discussed about the dataset has the most number of issue discussions (33\% of all issue discussions),  followed by discussions related to user opinions (20\% of issue discussions) and the feature engineering methodologies in the APIs (12\% of issue discussions). 
On the contrary, the least number of discussions in the issue discussions happened around topics like performance measurement (1\% of issue discussions) and troubleshooting (1\% of issue discussions).

\begin{figure}[t]
\centering
\includegraphics[scale=.5]{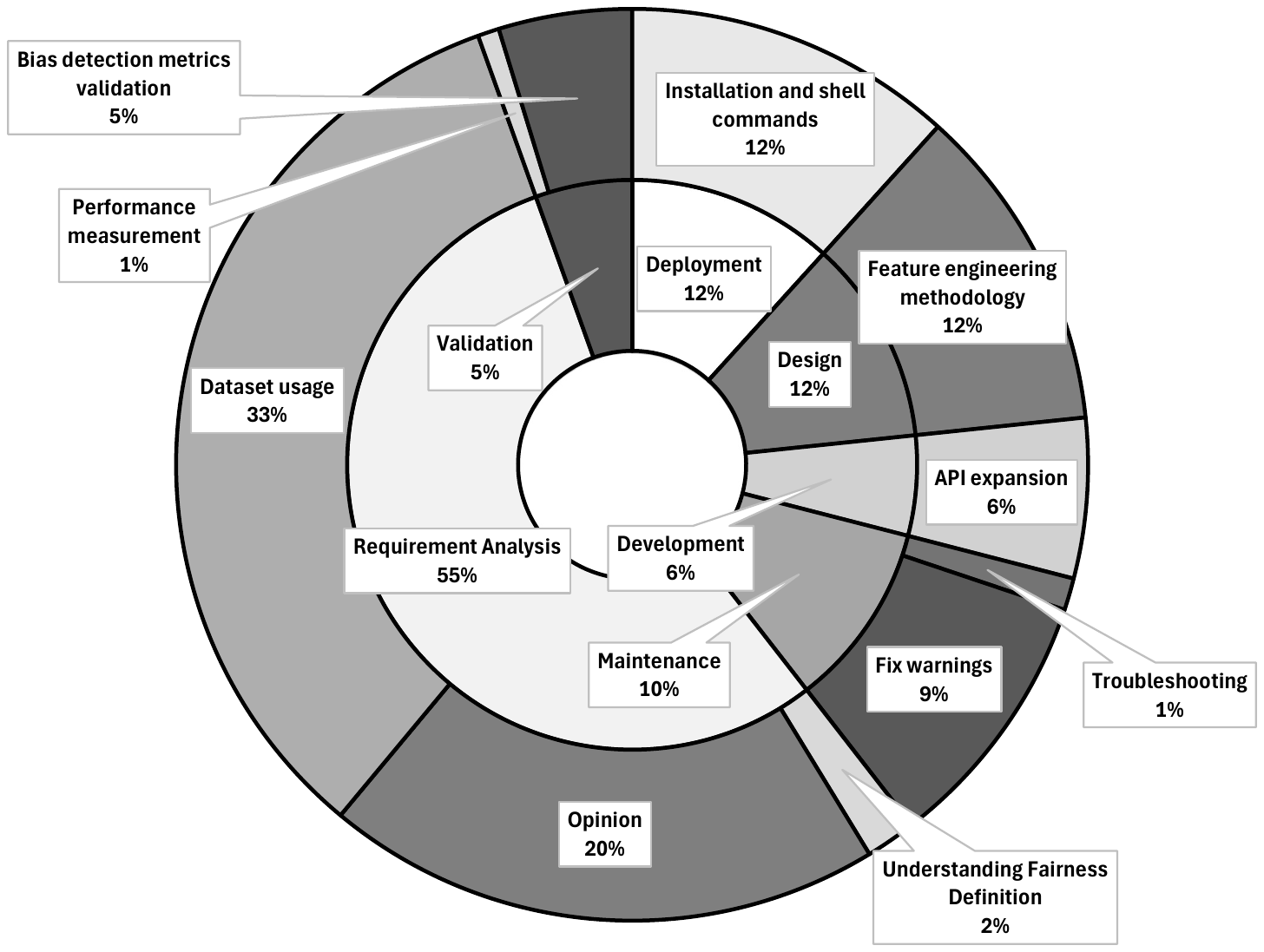}
\caption{Issue discussion topics with categories and subcategories}
\label{fig:topic_classify}
\end{figure}

Figure~\ref{fig:topic_classify} shows the taxonomy of the topics created by organizing the 10 topics into a hierarchy. As our categorization is based on the stages of SDLC, we have
six categories describing the different stages of the SDLC: Requirement analysis,
Design, Development, Deployment, Validation, and Maintenance. We describe these
categories in detail as follows.



\nd\bf{(1) Requirement Analysis Related Topics.}  This category has the highest coverage (55\% of discussions, 3 topics), which contains discussions related to analyzing the requirements of developing fair ML models. More specifically, it contains discussions or queries regarding how to use the bias detection and mitigation approaches correctly and efficiently. The three topics within this category are: 
\begin{enumerate}[(i)]
\item Dataset usage, which contains queries about the usage of datasets supported by the API libraries. For example, what to do about a certain dataset that throws an error (e.g., \href{https://github.com/fairlearn/fairlearn/issues/95#issuecomment-561857634}{fairlearn/issues/95\#issuecomment-561857634}).

\item Opinion, which contains queries related to asking opinions about fairness API-related problems. For example, it contains issues about if API developers should change the development process (e.g., \href{https://github.com/fairlearn/fairlearn/issues/853}{fairlearn/issues/853}), use certain external package as dependency or implement it (e.g., \href{https://github.com/fairlearn/fairlearn/issues/431#issuecomment-632558930}{fairlearn/issues/431\#issuecomment-632558930}).

\item Understanding fairness definition contains queries to understand the definition of fairness in certain contexts or to report issues regarding invalid or inconsistent use of fairness definitions. For example, suggesting to reconsider the use of the words \emph{bias} and \emph{prejudice}, in a library's README file (e.g., \href{https://github.com/Trusted-AI/AIF360/issues/97}{AIF360/issues/97}).
\end{enumerate}

\nd\bf{(2) Design Related Topics.} This category contains discussions or queries about the methodology of designing certain features utilizing the support provided by the API libraries. This category consists of topics related to Feature engineering methodologies, which contains queries related to applying feature engineering techniques or selecting features. For example, querying about how to use multiple sensitive features together (e.g., \href{https://github.com/fairlearn/fairlearn/issues/376}{fairlearn/issues/376})

\nd\bf{(3) Deployment Related Topics.} This category focuses on deployment-related issues that the developers and the users generally face while deploying the system. The topics covered under this category focus on Installation and shell commands. This contains discussions related to the installation of the libraries or their dependencies, and shell commands that can be used as an alternative means to access the API features as a command-line utility. For example, this topic contains a discussion about installation issues on certain operating systems (\href{https://github.com/Trusted-AI/AIF360/issues/44}{AIF360/issues/44}).


\nd\bf{(4) Maintenance Related Topics.} This is the fourth-largest category (10\% of the discussions) and consists of three topics that primarily focus on maintaining the source code of the API libraries. The two topics that are under this category are:
\begin{inparaenum}[(i)]
\item Troubleshooting, which contains bug reporting and bug fixing discussions. 
For example, reporting storage-related bugs while using an API (\href{https://github.com/fairlearn/fairlearn/issues/322}{fairlearn/issues/322}).
\item Fix warnings topic contains discussions on fixing warnings that can be generated in various phases of 
using the API library or the documentation. For example, a request to fix warnings in the documentation building phase (\href{https://github.com/fairlearn/fairlearn/issues/852}{fairlearn/issues/852}) is under this topic.
\end{inparaenum}

\nd\bf{(5) Development Related Topics.} Development consists of discussion around developing ML models or implementing fairness APIs.
This category covers the topics related to API extension. API expansion contains discussion around feature requests to expand the list of supported APIs. For example, asking to add a new fairness-aware algorithm as a new API  (e.g., \href{https://github.com/fairlearn/fairlearn/issues/956}{fairlearn/issues/956}), and asking to implement new sets of fairness metrics  (e.g., \href{https://github.com/dssg/aequitas/issues/48}{aequitas/issues/48}) are under this topic.

\nd\bf{(6) Validation Related Topics.}  This category contains discussions related to the validation of the developed fair ML model. The two categories that fall under this topic are:
\begin{inparaenum}[(i)]
\item Bias detection metrics validation, which contains queries regarding how to validate which bias detection metrics to apply in a certain scenario. For example, the question about how to differentiate between two bias detection metrics (\href{https://github.com/fairlearn/fairlearn/issues/466#issuecomment-649096969}{fairlearn/issues/466\#issuecomment-649096969}) is within this topic.
\item Performance measurement contains issues related to performance. For example, this topic contains the issue about the limitation of an existing performance measurement API (\href{https://github.com/fairlearn/fairlearn/issues/756}{fairlearn/issues/756}).
\end{inparaenum}

\begin{rqbox}{RQ4 Summary} 

We found 10 topics in the issue discussions of the 13 fairness API libraries. Our analysis reveals 
10 different kinds of issues that developers face while developing and using fairness libraries. 
These 10 topics are from all six phases of the software development life cycle.
We observe that developers frequently face significant
methodological, troubleshooting, and installation issues while developing and
using fairness libraries. The issues cover all the stages of SDLC, although most
of the issues are related to Requirements, Deployment, and Design.
Users need proper training on understanding what fairness is (especially in
a given context), what methodology to adopt, and how
to deal with these libraries. Our analysis, however, is limited to the issues posted and discussed within the libraries' repositories only. We could not capture issues discussed outside these repositories; thus, we may have missed other development challenges associated with these libraries.

\end{rqbox}

\section{Implications of Findings}\label{sec:discussions}
Our findings can be instrumental for the following stakeholders:
\begin{inparaenum}
\item \bf{Fairness API Developers}  like the authors of the fairness API libraries can further improve their APIs by addressing the common challenges we reported in the previous section.
\item \bf{ML Practitioners} can learn from the reported use cases where fairness APIs are used, to help their adoption of the APIs in various domains. 
\item \bf{ML researchers} can identify areas and domains where fairness APIs need to be adopted more (e.g., by identifying domains that are missing in our study).
\item \bf{SE researchers} can develop techniques to improve the quality assurance of the application of fairness APIs in ML software systems (e.g., by improving the troubleshooting process of the APIs).
\item \bf{ML Educators} can develop documentation and tutorials to improve the adoption of the fairness APIs, given that many repositories we observed are related to simply learning about the fairness APIs. 

\end{inparaenum} We further discuss the implications below.

\begin{figure}[t]
\centering
\includegraphics[scale=.5]{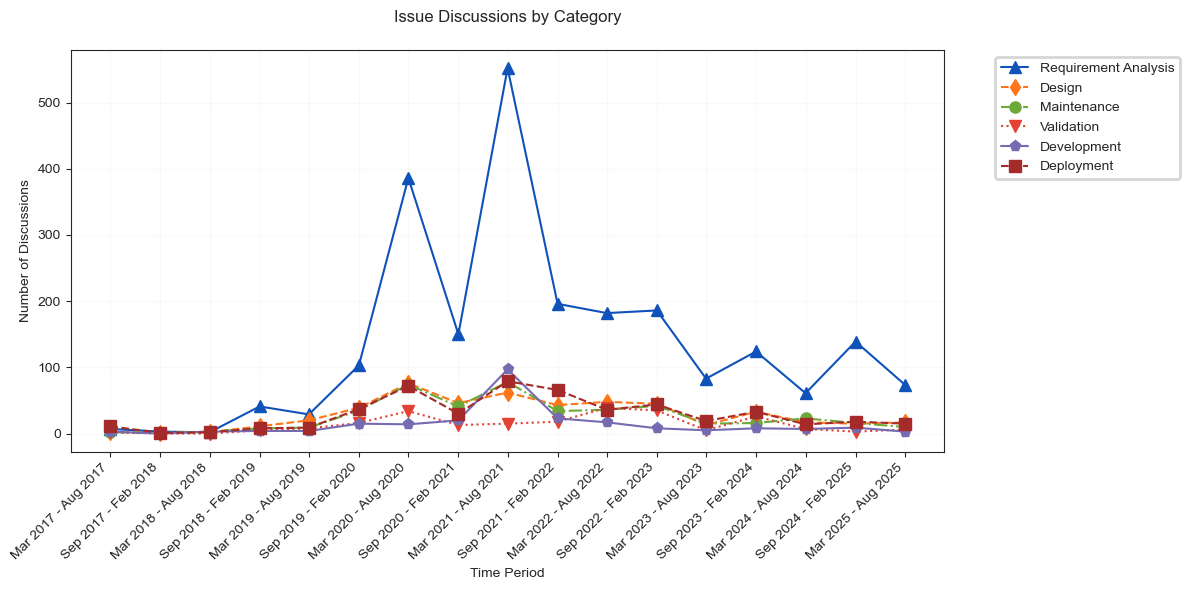}
\caption{The absolute number of issue discussions on different SDLC stages over time.}
\label{fig:topic_absolute_impact}
\end{figure}

\bf{\ul{Fairness API Developers.}} 

We found that around 19.8\% of the issue discussions were about asking for opinions (Figure~\ref{fig:topic_percentage}). These discussions often included topics such as what fairness issues to look for, what fairness metrics to use in certain situations, etc. 
As these types of questions, along with ``what is fair'' are not a trivial question to answer, the API developers may need to train themselves in various ML concepts and fairness in ML in general, to better prepare themselves in helping out their respective API user community.

Moreover, the study provides insights to the new fairness API developers about the challenges that their users may face in the future. \it{ In our study, a significant number of issues were related to API installation, release, and dependency, which implies the importance of providing usable documentation while releasing these fairness libraries.}

In \fig\ref{fig:topic_absolute_impact}, we show the evolution of the six SDLC phases across the 10 topics that 
we observed. Recall from \sec\ref{sec:challenges-bias-apis} that the six SDLC phases are used to group the 10 topics into higher categories in \fig\ref{fig:topic_classify}. 

We compute the absolute impact of each SDLC phase following Wan et al.~\cite{wanWhatProgrammersDiscuss2021}. First, we compute the number of new issues reported under each topic every six months in our API issue dataset from 
\sec\ref{sec:collect-issue-comments}. Second, we compute the number of new issues reported every six months for each SDLC phase by summing the total number of new issues reported under the topics assigned to that SDLC phase. The absolute impact for an SDLC phase is thus defined as the total number of new issues reported for that phase in each six-month interval.

 As shown in \fig\ref{fig:topic_absolute_impact}, the discussions about the SDLC phase ``Requirement Analysis`` are the most frequent over the years. 
We observed that this prevalence may be because fairness API developers often struggle to design features that meet requirements across diverse use cases.
As a result, many issue comments contain concrete opinions and suggestions from both API users and developers. These comments effectively serve as a collaborative venue of requirement collection. \it{To act on these findings, fairness API developers can systematically mine and categorize requirements from the issue comments of their project and from other related fairness APIs to identify the most recurrent needs.} \it{This could also open up 
a collaborative environment among the fairness API developers and a shared venue where they can discuss with each other the emerging requirements for their API libraries and whether and how 
they can work together to design their API libraries better for reuse.}

\begin{figure}[t]
\centering
\includegraphics[scale=.5]{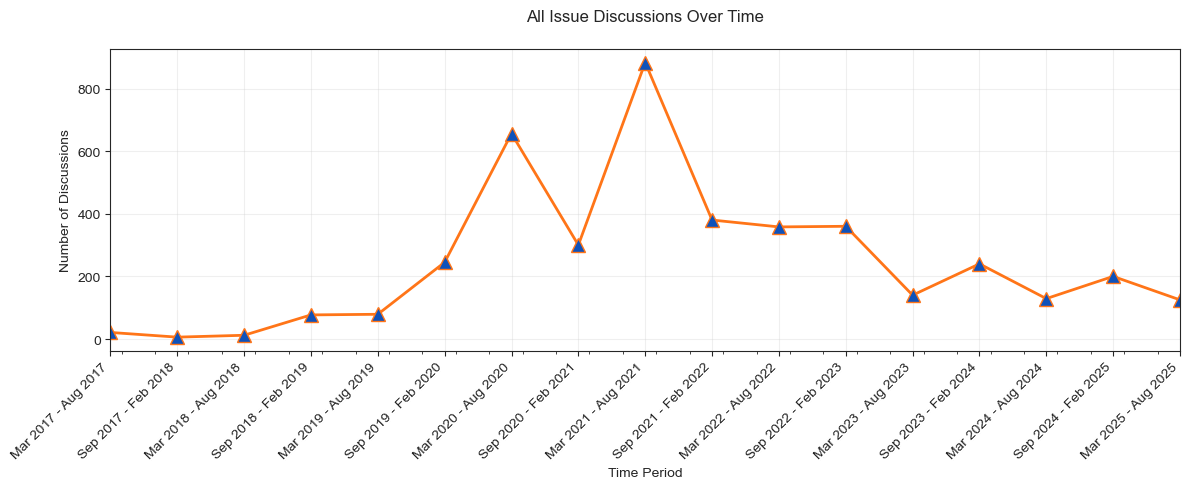}
\caption{Total absolute impact of the topic popularity.}
\label{fig:topic_discussion_by_year}
\end{figure}

\bf{\ul{ML Practitioners.}} In \fig\ref{fig:topic_discussion_by_year}, we show the total absolute impact of all topics across all the issue comments in our dataset. 
The line for total absolute impact denotes the total number of new issue comments reported per six-month period across all the fairness API libraries we studied. The line shows that 
there is a sharp increase in the total number of new issues reported from  August, 2019. One reason for this is that two major versions of the popular Fairlearn API 
were released during late 2019 and 2020. Another reason is that overall, all the 13 fairness API libraries we studied have experienced an increase in the number of issues reported in their corresponding issue tracking system on GitHub. 
The sharp increase in the number of reported issues is also attributed to the 
increase in the number of users logging those issues. The number of fairness APIs 
is low, and our observation in RQ1 (Section \ref{sec:lib_usecases}) also finds that the use cases can vary in terms of the usage of the APIs.
\it{As such, ML practitioners like data scientists can benefit from such insights to stay aware of the 
recent and emerging trends in the responsible ML area, which then can guide their learning of the APIs and their adoption into practical ML systems}. 

During their adoption of a fairness API, the ML practitioners can also benefit from the knowledge of how the APIs are currently being used to support 
diverse real-world scenarios. In \fig\ref{fig:usecase_categ_generic} and \ref{fig:usecase_categ_nongeneric}, we presented 17 unique use cases we observed in the GitHub ML software repos that reused our 13 fairness API libraries. 
\it{ Within the non-generic use cases, we see the applications related to various topics- Business (e.g., Credit Approval), Legal (e.g., Risk Detection), and Health (e.g., Patient Retention).
Such insights and code examples can inform ML practitioners like data scientists and ML policy researchers (e.g., national policymakers on responsible AI adoption) of the 
current state of the adoption and prevalence of fairness API developers - which then can guide them in their respective adoption of the APIs (e.g., for data scientists) as well as to 
formulate policies regarding the safe usage and adoption of the APIs across the diverse use cases (ML policymakers).}         

\bf{\ul{ML Researchers.}} Our analysis of the ML software repositories on GitHub shows 17 distinct use cases where fairness APIs are used. The diversity of the use cases 
shows that the design and development of bias detection and mitigation approaches need to incorporate concepts from multiple domains while ensuring that the developed approaches 
can be applied across diverse domains and use cases. \it{ML researchers can take note of our taxonomies of use cases (see \fig\ref{fig:usecase_categ_nongeneric}) during their design and improvement 
of bias detection and mitigation techniques.} At the same time, research in bias detection techniques finds that there can be too many bias detection metrics that are currently being 
proposed and it is not trivial to determine which metrics are more suitable for a given use case or domain (see `The Zoo of Fairness Metrics in Machine Learning' by Castelnovo et al. \cite{castelnovoZooFairnessMetrics2021}). 
Our investigation of the application of the bias detection metrics in \sec\ref{sec:rq-bias-detection-strategies} shows that not all the proposed metrics are used uniformly. 
For example, in \fig\ref{fig:detection_metric_types}, we show that 83\% of the bias detection APIs usage is done by the utility metrics. Recall from \sec\ref{sec:background-bias-detection} that 
the utility metrics for bias detection are used to aid the bias detection in any given scenario. Such metrics are mostly generic like the number of false positives or false negatives. 
\it{Our observation in \fig\ref{fig:detection_metric_types} thus can be utilized by ML researchers during their design of new bias detection algorithms like 
the use of utility metrics to ensure the easier adoption and usage of the metrics, because otherwise, users may find it difficult to understand a non-trivial metric and to explain its effectiveness 
within a given use case scenario.} 

Besides the utility metrics used for bias detection, significant research efforts have so far been devoted to developing specific metrics 
like Group, Individual, Group-Individual, and Subgroup (ref \sec\ref{sec:background-bias-detection} for an overview of the metric types). 
In \fig\ref{fig:detection-trends}, we show the evolution of the usage of the different bias detection metric types in our studied GitHub ML software repos. 
Among the non-utility metrics, Group fairness metrics are increasingly being used over time, while the usage of the other non-utility metrics (e.g., Individual) have been non-existent.
\it{This observation can guide the ML researchers during their design of new bias detection algorithms like 
focusing more on the metrics that can support Group fairness.}  
  
Once bias is detected in a dataset or a model, we can utilize the bias mitigation approaches to fix the dataset or model. Like the bias detection approaches, 
diverse techniques are developed to mitigate bias. As such, it is necessary to know whether all such mitigation techniques are used, because otherwise, 
the ML researchers might be spending time developing a mitigation technique that may not have many real-world values.
In \fig\ref{fig:mitigation_approach}, we show a taxonomy of the usage of bias mitigation techniques as we observed in our studied GitHub ML software. 
\it{As we can see, in-processing bias mitigation techniques are used in about 45\% of the observed scenarios, while techniques related to pre-processing account for around 37\% of the observed scenarios. ML researchers can use this finding to further focus on developing in-processing techniques that can assess and fix fairness constraints during ML model development, as in-processing techniques such as Fairness constraints accounted for 20.7\% of all mitigation scenarios. 
Moreover, ML researchers can use our findings to also focus on developing pre-processing techniques that can automatically fix biases within a training dataset by applying techniques such as Data transformation (accounted for 17.2\% of all mitigation scenarios in our dataset).}

We observed in our study that developers mainly focused on group fairness, with little to no adoption of subgroup or individual fairness approaches. This could be due to higher complexity associated with computing the other metrics, or the lack of supporting tooling, or insufficient guidance with regards to their practical benefits. ML researchers can address this lack of adoption of other fairness metrics by creating independent implementations, by providing better visualizations of subgroup/individual bias, and by conducting practical case studies that demonstrate their added value.

Furthermore, several mitigation approaches were never used by the developers in our studied projects, such as advanced data transformation techniques (e.g., adversarially learned fair representations), resampling methods (e.g., massaging, upsampling), certain fairness constraints (e.g., CORELS), and strategies like random flipping. Some of these are relatively new, while others (e.g., random flipping) may be seen as too simplistic to address fairness effectively. Researchers can investigate whether these methods are underused due to complexity, lack of documentation, or perceived ineffectiveness, and then either improve their usability or focus efforts on other strategies.

\begin{figure}[thb]
\centering
\includegraphics[scale=.5]{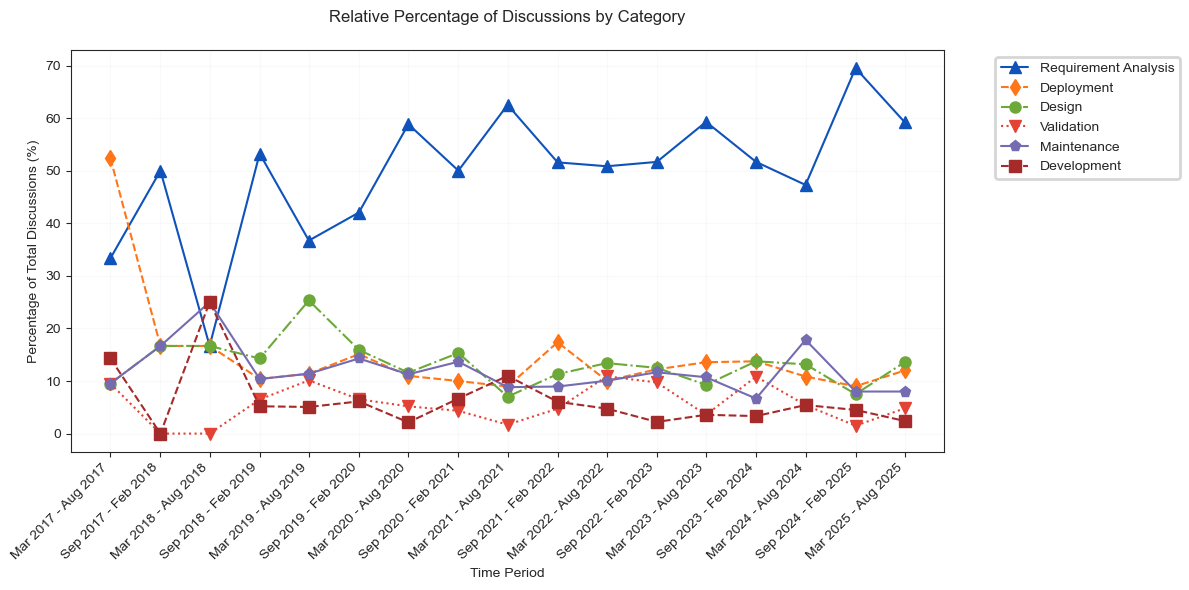} 
\caption{The relative number of issue discussions on different SDLC stages over time.}
\label{fig:topic_relative_impact}
\end{figure}
\bf{\ul{SE Researchers.}} Research in trustworthy ML software has so far focused on the 
quality assurance of supervised and deep ML algorithms, e.g., by checking for faults in the underlying software~\cite{nikanjamFaultsDeepReinforcement2021}.
We are not aware of research that focused on the quality assurance of ML software that utilizes the fairness APIs to detect and mitigate bias. 
Our observation of the topics in the issue comments of the 13 fairness API libraries shows that 
Requirement-related topics were the most dominant from the beginning of the data
collection.
\fig\ref{fig:topic_absolute_impact} shows the number of discussions over time
for each category. Requirement-related topics have always been the most dominant. 
Deployment, Design, and Maintenance have been the three other most
discussed categories, which is consistent with the summarized values presented
in \fig\ref{fig:topic_classify}.
\fig\ref{fig:topic_relative_impact} compares these six categories in relative
values, where we calculated the percent of discussions under each category. The
observations are quite similar to the absolute values
(\fig\ref{fig:topic_relative_impact}). Evidently, our summarized observations
(\fig\ref{fig:topic_classify}) are very similar to our time-variant
observations. We conclude that the issues and challenges that developers face
while developing and using fairness libraries remain the same; they have not
changed in the last several years. \it{Intuitively, a fairness API is mainly designed and developed by ML researchers. As such, the requirement 
analysis and design of the API and the algorithms become the major source of discussion in the issue-tracking systems of the API libraries. 
However, given the complexity of the adoption of such fairness APIs in real-world ML software, it is important that SE researchers join 
hands with the ML researchers to incorporate quality assurance techniques from SE into the fairness APIs. With such increased collaboration, 
we may experience more discussions around the deployment and validation of the APIs. This should be encouraging
for the research community for producing better and more robust
fairness algorithms.}

In our studied ML software repositories, we did not observe any use case to detect or mitigate bias in SE scenarios.
However, the challenges that the API developers are facing could potentially encourage ML
researchers to further explore the issues for suggesting solutions to minimize
their effects. For many research areas---e.g., defect
prediction~\cite{tantithamthavornExperienceReportDefect2018}, fault localization~\cite{kochharPractitionersExpectationsAutomated2016},
and software energy efficiency~\cite{manotasEmpiricalStudyPractitioners2016, pangWhatProgrammersKnow2016} --- there exists significant  gaps between practice and research \cite{tantithamthavornExperienceReportDefect2018, kochharPractitionersExpectationsAutomated2016, manotasEmpiricalStudyPractitioners2016, pangWhatProgrammersKnow2016}. 
Several research efforts in SE find that bias can exist in 
datasets prepared for fault localization or during the measurement of certain metrics \cite{shepperdResearcherBiasUse2014,rahmanSampleSizeVs2013,yaoImpactUsingBiased2021}.
The ML developers and SE researchers could work together to leverage state-of-the-art research about bias detection and mitigation to achieve fairness in the domain of SE. 

\bf{\ul{ML Educators.}} Our results imply the importance of a more up-to-date curriculum for 
future ML practitioners that the educators should focus on. We observed a
significant portion of the issue discussions are related to asking for opinions and
querying about how to use the APIs and datasets properly (see \fig\ref{fig:topic_percentage}). Therefore, educators
should focus on defining \textit{context-based fairness} and should develop
strategies for achieving them.  Also, more than half of the bias API usage
in our dataset is related to generic learning-focused use-cases (see \fig\ref{fig:usecase_categ_generic}). In addition, we observed a lot of troubleshooting, and installation problems in our issue discussion dataset (see \fig\ref{fig:topic_classify}).
This implies that the educators should facilitate the hands-on experience for the
developers in building and using fairness libraries. These observations present a call to arms to the educators for 
developing more state-of-the-art curricula, which train practitioners in dealing with fairness libraries at all the stages of the software development life cycle.

\section{Threats to Validity}\label{sec:threats}

\begin{table*}[t]
    \centering
    \caption{Issue discussion topics and related number of libraries addressing each}
    \label{tab:topic_distribution}
    \rowcolors{1}{}{lightgray!30}
    \scalebox{1.0}{
        \begin{tabular}{lc}
            \toprule
            \textbf{Topic Name} & \textbf{\# Number of Involved Libraries} \\ 
            \midrule
            API Expansion                         & 9  \\
            Bias Detection Metrics Validation     & 6  \\
            Dataset Usage                         & 11 \\
            Feature Engineering Methodology       & 10 \\
            Fix Warnings                          & 8  \\
            Installation and Shell Commands       & 9  \\
            Opinion                               & 9  \\
            Performance Measurement               & 10 \\
            Troubleshooting                       & 8  \\
            Understanding Fairness Definition     & 8  \\
            \bottomrule
        \end{tabular}
    }
\end{table*}

\bf{\ul{External Validity}} concerns about the generalizability of 
findings. We only used open-source GitHub projects to collect our repository
dataset. We do not know if our findings would be consistent with closed-source
projects. Also, searching for repositories using GitHub API has its own
limitation. This API only provides repositories that were active within the
current year. Hence we might have missed some repositories that were not active
recently. Moreover, we could only get up to 3000 repositories per library even
if the search query matches with more than 3000 repositories.  When searching
for library imports within the studied repositories, we have considered only
Python (\texttt{.py}) source files. In this approach, we might have missed some
library usage that exists within files with other formats (such as Jupyter
notebook's \texttt{.ipynb} files).

In our issue discussion dataset, the API library Fairlearn alone accounts for more than half of the total discussions. This is partly due to Fairlearn being one of the most popular libraries in our list. While it is possible to reduce its influence by excluding it or sampling only a subset of its issues, doing so would significantly reduce the size of our dataset. This is a concern given that our final dataset comprises only 13 libraries after applying a rigorous selection process. Therefore, we chose to retain all available data to preserve statistical power. Nevertheless, it is important to acknowledge that a large number of samples from a single repository may introduce bias into the analysis. Nevertheless, we acknowledge that such an imbalance may introduce bias into the analysis. To evaluate this concern, we examined the distribution of discussed topics across libraries. As shown in Table \ref{tab:topic_distribution}, each topic is represented by issues from at least six different libraries, with most topics involving eight or more. This demonstrates that, despite Fairlearn's dominance in volume, the thematic diversity of the dataset is preserved. Thus, our findings reflect a broad range of perspectives and are not driven solely by a single repository.


\nd\bf{\ul{Internal Validity}} concerns about the degree to relationship between different variables in the study. We used Abstract Syntax Tree (AST) analysis for locating the library usages. Static AST analysis, however, can not handle all possible cases (e.g., wildcard imports, and imports using specific path syntax). Moreover, static AST analysis has limitations regarding detecting runtime library imports or API calls.

Most of the findings of our study are heavily based on manual labeling---e.g., labeling of the API use-cases and the issue discussion topics. We also used manual verification to filter out irrelevant repositories from the repository dataset. It is difficult to guarantee that identical results will be produced with a replication study. These threats, however, were mitigated to some extent because multiple authors have worked together, and decisions were made based on agreements. 

Additionally, we had to provide the number of topics in advance to tell the LDA algorithm how many topics we wanted. Estimating the number of optimal topics can be inaccurate without several experiments. We, therefore, mitigated this threat by repeating our experiments multiple times for finding the optimal coherency score. This is a standard approach followed by early LDA-related studies (e.g., \cite{uddinAutomaticSummarizationAPI2017,uddinEmpiricalStudyIoT2021,alaminEmpiricalStudyDeveloper2021}). In addition to using the coherence score, we also manually examined the generated topics to verify their relevance. However, we must admit that the accuracy of manual labeling by human participants remains a threat to any LDA-related studies.

We included all of the issues and issue comments we found from the 11 API library repositories. Different filtering criteria could have been applied during our selection of issue discussions. However, we did not want to exclude any such comments, whether short or long, because we wanted to see all the possible issues that were faced by the developers. This should not impact the overall outcome because after topic modeling and topic labeling steps interesting issues such as metrics operations, and bias detection and mitigation issues have resulted in separate categories than simple installation problems.

For performing the trend analysis presented in RQ2 and RQ3 (see \sec\ref{sec:rq-bias-detection-results} and \ref{sec:rq-bias-mitigation-results} respectively), we needed to collect the full git source tree for the repositories. However, during the time of these analyses (Oct 01, 2022) we could not collect the full source code of 6 out of 84 repositories (due to issues related to cloning failure and removal of some repositories from GitHub). As a result, for performing the trend analysis we used 78 repositories instead. This, however, should not significantly impact our presented trend analyses as we had still about 93\% of the repositories left for performing the analyses.

\nd\bf{\ul{Construct Validity}} is related to the potential errors that may have occurred when preparing the API library list and choosing API code search to construct the repository dataset. Also, for understanding the challenges that the developers face (\textbf{RQ4}), we have relied only on the GitHub issues. It is possible that the developers have also reported and discussed the issues to other external issue-reporting platforms (e.g., Bugzilla, and JIRA). However, generally, if the developers use external sites for such discussions, the API library's GitHub source code repositories or their respective websites mention those details so that the users of these libraries may be able to report the issues there. In this study, we have not found any such mention of JIRA/Bugzilla in their GitHub readmes or websites.

Fairness is a relatively new concept in ML. As such, the adoption of fairness APIs is not 
as widespread as traditional ML APIs. Therefore, our list of obtained repositories is small, and it probably 
offers little insights to analyze those repositories using
popularity-based (e.g., star count, fork count) or development activity-based
(e.g., commit count, developer count) filtering criteria. This is because such filtering would 
further reduce the number of repositories. Additionally, in the later stages of our analysis, we found out
that a good portion of the selected repositories has focused on generic use cases that are less popular and had fewer development
activities. Our filtering based on popularity and development activities would
have removed such repositories and resulted in a very limited number of
repositories to perform the rest of our study. However, we divided our analysis in RQ1-RQ3 by isolating the non-generic use 
cases from generic ones, to highlight how fairness APIs are used in real-world applications.

\section{Related Work}\label{sec:related_work}

In this section, we first discuss the studies that had been performed investigating the impact of bias in various ML software systems. Then we discuss the existing approaches proposed by the research community to ease the task of identifying and mitigating bias in ML models. Finally, we finish by describing the existing contributions of the software engineering research community to aid in this process. 

\subsection{Impact of bias in ML software systems}
Bias in ML software systems can affect our life deeply as more and more ML systems are being deployed in making real-life decisions. The biased behaviors of these ML systems can make life-changing decisions that ultimately can have a deep impact on the whole society. For example, Amazon's use of a hiring system that was biased toward women has impacted the hiring process of the company~\cite{AmazonScrapsSecret}. A study conducted by Angwin et al.~\cite{mattujefflarsonMachineBias} showed the effect of the use of the ML system in making biased life-impacting decisions toward black defenders as the system labeled black defenders as high risk of being re-offender.

Being a matter of high importance, the research community was not silent regarding the analysis of bias and there have been significant studies in identifying and mitigating bias across various domains. Some of such studies which have focused on a variety of domains are as follows:
\begin{inparaenum}[(a)]
\item Schwemmer et al.~\cite{schwemmerDiagnosingGenderBias2020} evaluated gender bias in commercial image recognition systems and found that even though the performance of these engines was pretty good, the outcomes were biased towards women.
\item Saumya Bhadani~\cite{bhadaniBiasesRecommendationSystem2021} investigated how popularity bias impacts the journalistic quality in online news websites and proposed a method to quantify this bias using empirical analysis.
\item Hovy and Prabhumoye~\cite{hovyFiveSourcesBias2021} discussed five sources of bias in natural language processing systems and the countermeasures that can be applied to minimize them.
\item Vokinger et al.~\cite{vokingerMitigatingBiasMachine2021} discussed solutions that can be applied in different development stages of the ML model development pipeline to mitigate bias in ML systems that are used in machines and potentially impact patients' clinical care.
\end{inparaenum}


Even though bias in an ML system can be introduced in various ways, in most cases, bias comes from utilizing a biased dataset in training the ML models. There also exist various studies that describe the various ways through which bias can be introduced into the ML systems~\cite{mehrabiSurveyBiasFairness2021,baeza-yatesBiasWeb2018,garrido-munozSurveyBiasDeep2021}. For example, Mehrabi et al.~\cite{mehrabiSurveyBiasFairness2021} discussed how 19 different kinds of bias might be introduced in the ML systems through three different kinds of interaction. 

Even though our study does not directly study the impact of bias in any specific ML domains, we report the use cases and domains in which we see the users use bias detection and mitigation APIs to analyze and minimize these impacts of unfairness.  

\subsection{Bias detection and mitigation approaches}
Even though biases have been found in various domains and various ML systems, defining bias in a generic way is not a straightforward task. There have been various attempts to define and detect bias utilizing various statistical and probabilistic approaches. Researchers came up with various metrics such as disparate impact~\cite{feldmanCertifyingRemovingDisparate2015}, statistical parity \cite{dworkFairnessAwareness2011}, equality of opportunity \cite{hardtEqualityOpportunitySupervised2016a}, etc. to measure bias in different scenarios. Along with defining these metrics, studies have also explored algorithms to detect bias in ML systems~\cite{hardtEqualityOpportunitySupervised2016a,gajaneFormalizingFairnessPrediction2018,vermaFairnessDefinitionsExplained2018,yaoParityFairnessObjectives2017,pessachAlgorithmicFairness2020}.

Mitigating bias in the ML system is the primary way to ensure fair treatment for all of the involved stakeholders. So, the research community has been proposing various types of algorithms to minimize bias in a system and improve the overall fairness in ML systems' decision-making process~\cite{feldmanCertifyingRemovingDisparate2015,kamishimaFairnessawareClassifierPrejudice2012,zhangMitigatingUnwantedBiases2018,calmonOptimizedPreProcessingDiscrimination2017}. Approaches such as Adversarial debiasing~\cite{zhangMitigatingUnwantedBiases2018}, Disparate impact remover~\cite{feldmanCertifyingRemovingDisparate2015}, Prejudice remover~\cite{kamishimaFairnessawareClassifierPrejudice2012}, Reweighing~\cite{kamiranDataPreprocessingTechniques2012a}, etc. are some of the notable bias mitigation algorithms that are being used in the wild. 

Even though in this study we are not proposing any novel approach to identify or mitigate bias, we report how these existing algorithms from the literature have been implemented as APIs and investigate how they are being used across various domains and use cases.

\subsection{Bias-related studies in Software Engineering}
\subsubsection{Investigative studies}
Studies have also been performed to explore bias in software engineering settings. For example, Brun and Meliou discussed how bias may exist in ML software development pipeline \cite{brunSoftwareFairness2018}. Moreover, there have been studies to investigate gender bias in software engineering-based sentiment analysis systems \cite{imtiazInvestigatingEffectsGender2019,paulExpressionsSentimentsCode2019}. Additionally, several studies have also been performed on exploring bias in software defect prediction-focused ML systems \cite{shepperdResearcherBiasUse2014, rahmanSampleSizeVs2013, tantithamthavornCommentsResearcherBias2016, shepperdAuthorsReplyComments2018}. For example,  Shepperd et al.~\cite{shepperdResearcherBiasUse2014} investigated researcher bias in the use of ML in software defect prediction, and Rahman et al.~\cite{rahmanSampleSizeVs2013} investigated the effect of sample size in contrast to bias in defect prediction datasets. 

None of these studies investigates what challenges the fairness API developers usually face while developing these fairness APIs or how these APIs are being used in the wild. 
Our study adds value to the existing literature by qualitatively investigating into these two unexplored topics.

\subsubsection{Bias detection and mitigation tools} The software engineering research community has also developed automated tools to help ML developers in identifying and mitigating bias.  For example, Chakraborty et al. proposed multiple approaches such as Fairway~\cite{chakrabortyFairwayWayBuild2020}, Fair-Smote~\cite{chakrabortyBiasMachineLearning2021}, Fair-SSL~\cite{chakrabortyFairSSLBuildingFair2022}, etc. as improvements over the state-of-the-art algorithms. Galhotra et al.~\cite{galhotraFairnessTestingTesting2017} proposed Themis, a fairness testing approach, focusing on detecting causal discrimination which can generate discrimination tests automatically without requiring an oracle. Peng et al.~\cite{pengXFAIRBetterFairness2022} proposed a model-based extrapolation method called xFair that can be used in both improving fairness and explaining the cause of unfair behavior of a system. Yu et al.~\cite{yuFairBalanceImprovingMachine2021} worked on improving ML fairness on multiple sensitive attributes by utilizing the data balancing technique. Aggarwal et al.~\cite{aggarwalBlackBoxFairness2019} proposed a methodology for generating testing input automatically in order to identify individual discrimination utilizing blackbox testing mechanism. Yang et al.~\cite{yangBiasRVUncoveringBiased2021} proposed a tool called BiasRV that can notify the system admin when a deployed sentiment analysis system makes a biased prediction on a given input text. Asyrofi et al.~\cite{asyrofiBiasFinderMetamorphicTest2021} presented a tool called BiasFinder that aims to discover biased predictions in the sentiment analysis system utilizing the metamorphic testing approach.


\subsubsection{Empirical Studies} There also have been several empirical studies investigating the state and impact of bias in overall ML development and user community. For example, Harrison et al.~\cite{harrisonEmpiricalStudyPerceived2020} conducted an empirical study utilizing 502 Mechanical Turk workers on the perceived fairness of realistic and imperfect ML models. Biswas and Rajan~\cite{biswasMachineLearningModels2020} performed an empirical study on 40 ML models collected from Kaggle and evaluated the state of fairness using a comprehensive set of fairness metrics. Biswas and Rajan~\cite{biswasFairPreprocessingUnderstanding2021} also performed a study to understand the fairness impact of the ML pipelines' preprocessing stages by introducing the casual method of fairness and found that certain data transformer in ML pipeline causes the models to exhibit unfair behavior during this empirical study.

All of these empirical works, however, primarily focused either on specific ML models or were too limited by their selection of the ML development platforms from where they collected these models. None of these studies neither focused on analyzing the ecosystem of ML software repositories nor the usage of bias detection or mitigation APIs. 
So, in this study, we extend the existing literature by performing an empirical study on a large dataset of 230 ML  software repositories and study the usage of bias detection and mitigation APIs from 13 commonly used bias-related libraries in these repositories. We also analyze the challenges that the API developers usually face while implementing these API libraries.

\section{Conclusions}\label{sec:conclusions}
If bias in the ML system has not been taken care of, it can potentially be
devastating for the users of that ML system. Existing literature primarily
focuses on developing new metrics and algorithms to identify and mitigate biases
in different ML systems. But the success of these approaches depends on the
acceptance of the developers' communities that are developing real ML software
systems. In this paper, we have conducted a preliminary study to understand in what
scenarios fairness libraries are used in the real-world ML systems, and how the
developers ensure such fairness by detecting and mitigating relevant biases that
may exist in such systems. We have also investigated what frequent issues the
developers discuss about while dealing with fairness libraries. This study
reveals that fairness is a concern even in some very sensitive use-cases including
health, business, and legal decision-making ML systems (\textbf{RQ1}, \sec\ref{sec:lib_usecases}).   We
observed that bias detection is most common at higher level granularity, such as
group-level (\textbf{RQ2}, \sec\ref{sec:rq-bias-detection-strategies}). This is concerning because bias may exist at lower
level granularity, such as subgroup or individual levels, which will not be
mitigated by ensuring fairness at the higher levels only.  Among all the
mitigation approaches, in-processing is the most popular and it ensures
fairness within the ML pipeline/model itself (\textbf{RQ3}, \sec\ref{sec:rq-bias-mitigation-strategies}).  At the end, we
showed that ML developers are yet to master the art of understanding what
\textit{fairness} is in a given context, and how to deal with them
\textit{efficiently} (\textbf{RQ4}, \sec\ref{sec:challenges-bias-apis}). 
The findings from our study can guide several stakeholders including the fairness API developers, ML researchers, practitioners, and educators and 
SE researchers. 
Our future work will study more fairness API libraries and ML software repositories to find the 
root causes and the solution of the challenges that the API
developers are facing, to study the difficulties of these challenges, and to develop solutions to address the challenges.

\urlstyle{rm}
\bibliographystyle{abbrv}


\end{document}